\documentclass[sigconf, nonacm]{acmart}

\newcommand\vldbdoi{XX.XX/XXX.XX}

\newcommand\vldbavailabilityurl{https://github.com/TUDB-Labs/mLoRA}
\newcommand\vldbpagestyle{plain}

\renewcommand\footnotetextcopyrightpermission[1]{}

\usepackage{algorithm}
\usepackage{algpseudocode}
\usepackage{listings}
\usepackage{amsmath}
\usepackage{subfig}
\usepackage{pifont}
\usepackage{makecell}

\DeclareCaptionFormat{mylst}{\hrule#1#2#3}
\begin{document}
\title{mLoRA: Fine-Tuning LoRA Adapters via Highly-Efficient Pipeline Parallelism in Multiple GPUs}

\author{Zhengmao Ye}
\email{yezhengmaolove@gmail.com}
\authornote{These authors contributed equally to the paper}
\affiliation{%
   \institution{Sichuan University}
}

\author{Dengchun Li}
\email{mikecovlee@163.com}
\authornotemark[1]
\affiliation{%
   \institution{Sichuan University}
}

\author{Zetao Hu}
\email{vinkle-hzt@outlook.com}
\authornotemark[1]
\affiliation{%
   \institution{Sichuan University}
}
   
\author{Tingfeng Lan}
\email{tafflan2001@gmail.com}
\affiliation{%
   \institution{Sichuan University}
}

\author{Jian Sha}
\email{jian.sha@antgroup.com}
\affiliation{%
   \institution{Ant Group}
}

\author{Sicong Zhang}
\email{zsc@ix.cn}
\affiliation{%
   \institution{Zhejiang computation}
}

\author{Lei Duan}
\email{leiduan@scu.edu.cn}
\affiliation{%
   \institution{Sichuan University}
}
\author{Jie Zuo}
\email{zuojie@scu.edu.cn}
\affiliation{%
   \institution{Sichuan University}
}

\author{Hui Lu}
\email{hui.lu@uta.edu}
\affiliation{%
   \institution{The University of Texas at Arlington}
}

\author{Yuanchun Zhou}
\email{zyc@cnic.cn}
\affiliation{%
   \institution{CNIC, Chinese Academy of Science}
}

\author{Mingjie Tang}
\email{tangrock@gmail.com}
\affiliation{%
   \institution{Sichuan University}
}
   
\begin{abstract}
Transformer-based, pre-trained large language models (LLMs) have demonstrated outstanding performance across diverse domains, particularly in the emerging {\em pretrain-then-finetune} paradigm. Low-Rank Adaptation (LoRA), a parameter-efficient fine-tuning method, is commonly used to adapt a base LLM to multiple downstream tasks. Further, LLM platforms enable developers to fine-tune multiple models and develop various domain-specific applications simultaneously.
However, existing model parallelism schemes suffer from high communication overhead and inefficient GPU utilization when training multiple LoRA tasks across GPUs and machines.

In this paper, we present mLoRA, a parallelism-efficient fine-tuning system designed for training multiple LoRA across GPUs and machines. mLoRA introduces a novel LoRA-aware pipeline parallelism scheme that efficiently pipelines independent LoRA adapters and their distinct fine-tuning stages across GPUs and machines, along with a new LoRA-efficient operator to enhance GPU utilization during pipelined LoRA training. Our extensive evaluation shows that mLoRA can significantly reduce average fine-tuning task completion time, e.g., by 30\%, compared to state-of-the-art methods like FSDP. More importantly, mLoRA enables simultaneous fine-tuning of larger models, e.g., two Llama-2-13B models on four NVIDIA RTX A6000 48GB GPUs, which is not feasible for FSDP due to high memory requirements. Hence, mLoRA not only increases fine-tuning efficiency but also makes it more accessible on cost-effective GPUs. mLoRA has been deployed in AntGroup's production environment.
\end{abstract}

\maketitle

\pagestyle{\vldbpagestyle}


\section{Introduction}

Transformer-based, pre-trained large language models (LLMs), such as Gemma~\cite{gemma}, LLaMA~\cite{llama}, Mistral~\cite{mistral}, and Phi-3\cite{abdin2024phi3} have expanded their reach beyond natural language processing to a broad range of domain-specific tasks. This is achieved by adapting pre-trained LLMs for downstream tasks via {\em fine-tuning}, which enhances model performance for a particular task with brief training on task-specific data~\cite{finetune-1,finetune-2}. 
Examples of this adaptation include translating natural language questions into SQL queries for relational databases~\cite{texttosql}, converting heterogeneous data lakes into structured, queryable tables~\cite{arora2023language}, analyzing network traffic to enhance performance in network-related tasks~\cite{netgpt}, and others~\cite{time-llm, vision-language, kg-llm, FeuerLHF24, lao2024gptuner}.

As the size of LLMs grows exponentially -- rising from hundreds of billions to the anticipated trillions of model parameters~\cite{trillion} -- fine-tuning these models using traditional {\em full-weight} approaches, which require updating all parameters, becomes very expensive.
Instead, Parameter-Efficient Fine-Tuning (PEFT) methods~\cite{han2024parameterefficient}, including partial~\cite{ding2023parameter, zhao2020masking, ansell2021composable}, additive~\cite{rebuffi2017learning, houlsby2019parameter, lester2021power, asai2022attempt}, and reparameterized~\cite{lora} fine-tunings, have been developed. 
They train a much smaller set of parameters, thus cutting training costs while maintaining performance levels comparable to full-weight fine-tuning.

Low-Rank Adaptation (LoRA)~\cite{lora, dettmers2024qlora, chavan2023one}, a popular class of PEFT methods, freezes the parameters of an LLM while updating pairs of low-rank matrices with far fewer parameters, namely {\em adapter weights}. Models fine-tuned with LoRA not only match but also exceed the performance of fully fine-tuned models while remaining extremely lightweight, e.g., requiring less than 1\% of trainable parameters~\cite{han2024parameterefficient, lora-land}.
The cost-effectiveness and high performance of LoRA have spurred the development of numerous custom LLMs, each exhibiting notable performance in its specific domain~\cite{pycorrector, tianpeng, rs-llava}. It further facilitates scalable, large-scale serving platforms that can manage thousands of fine-tuned models on a single GPU~\cite{lorax} or across multiple GPUs~\cite{s-lora}. 

While recent attention has largely focused on LLM serving, such as resource efficiency, serving latency, scalability, scheduling, fairness, and multi-tenancy~\cite{xiaflash, kwon2023efficient, 280922, sheng2023fairness, s-lora, punica, wu2023fast, dlora}, less attention has been paid to addressing an equally important question: {\em how to effectively and efficiently build these fine-tuned variants?} Unlike training an LLM from scratch, which can require thousands of GPUs and days of time~\cite{jetmoe,vtrain}, lightweight LoRA enables a single GPU to build multiple model variants simultaneously, with even greater capacity when using multiple GPUs on one or multiple machines. Meanwhile, concurrently fine-tuning multiple adapters has become increasingly crucial: LLM platforms~\cite{llmplat1, llmplat2, llmplat3} enable developers to fine-tune multiple models and develop various domain-specific applications at the same time; for individual developers, selecting multiple sets of hyperparameters (e.g., learning rate or LoRA rank) either manually or automatically~\cite{tribes2023hyperparameter} by fine-tuning multiple adapters can quickly reveal the best-performing adapter.

However, the unique characteristics of LoRA present key challenges for parallel fine-tuning LoRA adapters. Conceivably, the frozen base LLM in LoRA facilitates the parallel training of multiple LoRA adapters by {\em sharing the same base model}, which reduces the GPU memory footprint (i.e., requiring only one copy of the LLM) and enhances training parallelism (i.e., allowing simultaneous LoRA training tasks). Nevertheless, when fine-tuning massive LoRA adapters exceeds the capacity of a single GPU, multiple GPUs become necessary; distributing a base model across GPUs involves {\em model parallelism}, which partitions the base model’s parameters and adapters and distributes them among these GPUs. Unfortunately, existing model parallelism approaches, such as tensor parallelism~\cite{megatron-lm,jia2019beyond} and pipeline parallelism~\cite{gpipe,fan2021dapple}, are plagued by {\em high communication overhead} due to the need for inter-GPU or inter-machine synchronization or {\em inefficient GPU utilization} caused by pipeline bubbles. Moreover, the small size of LoRA adapters exacerbates the issue -- training numerous small adapters in parallel results in frequent GPU kernel launches, which can substantially increase the total training time (e.g., up to 10\%).

To overcome these challenges, we present {\bf mLoRA}, a fine-tuning system designed and developed for efficiently fine-tuning LoRA adapters across multiple GPUs and machines. The key goal of mLoRA is to achieve high fine-tuning performance -- i.e., with low training latency and high training throughput -- by fully utilizing multi-GPU resources, including both computation and memory.

mLoRA first introduces a novel {\em pipeline parallelism} mechanism called LoRAPP, which ensures low communication overhead, high parallelism, and improved GPU efficiency for multi-LoRA, multi-GPU fine-tuning. LoRAPP capitalizes on the observation that although different LoRA adapters share the same base model, they can be trained independently without computational dependencies. This enables mLoRA to avoid multi-GPU fine-tuning pipeline stalls by freely and concurrently scheduling distinct training stages (e.g., forward and backward propagation) of different fine-tuning tasks, thus eliminating pipeline bubbles (i.e., {\em zero} bubbles). Further, mLoRA boosts GPU efficiency with a new {\em operator}, BatchLoRA. This operator consolidates multiple LoRA fine-tuning tasks into a large batch and performs collective matrix multiplication operations for all involved adapters rather than handling them individually. This approach enhances GPU utilization and reduces kernel launch overhead while maintaining model quality.

We have evaluated mLoRA by fine-tuning multiple LoRA adapters on various publically available LLMs of different sizes, e.g., TinyLlama-1.1B~\cite{tinyllama}, Llama-2-7B, and 13B~\cite{llama2}. Experiments demonstrate that mLoRA significantly reduces the completion time for fine-tuning tasks. For instance, it achieves a reduction in fine-tuning time by up to 45\% for the Llama-2-7B model in fp32 precision across four NVIDIA RTX A6000 48GB GPUs, compared to state-of-the-art methods like FSDP~\cite{fsdp}, which is an industry-grade parallel LLM training strategy.
Moreover, mLoRA enables the simultaneous fine-tuning of {\em larger} models, e.g., two Llama-2-13B models in fp32 precision with 4 NVIDIA RTX A6000 48GB GPUs, while FSDP cannot due to higher memory requirements. With its high fine-tuning efficiency and low cost, mLoRA addresses the critical issue of the scarcity and expense of high-end GPUs and has been deployed in the production environment at AntGroup, where it reduces the time for selecting optimal hyperparameters for LLM models by 30\%.

\begin{figure}[t]
    \centering
    \subfloat[\centering LoRA Approach]{{\includegraphics[width=.24\textwidth]{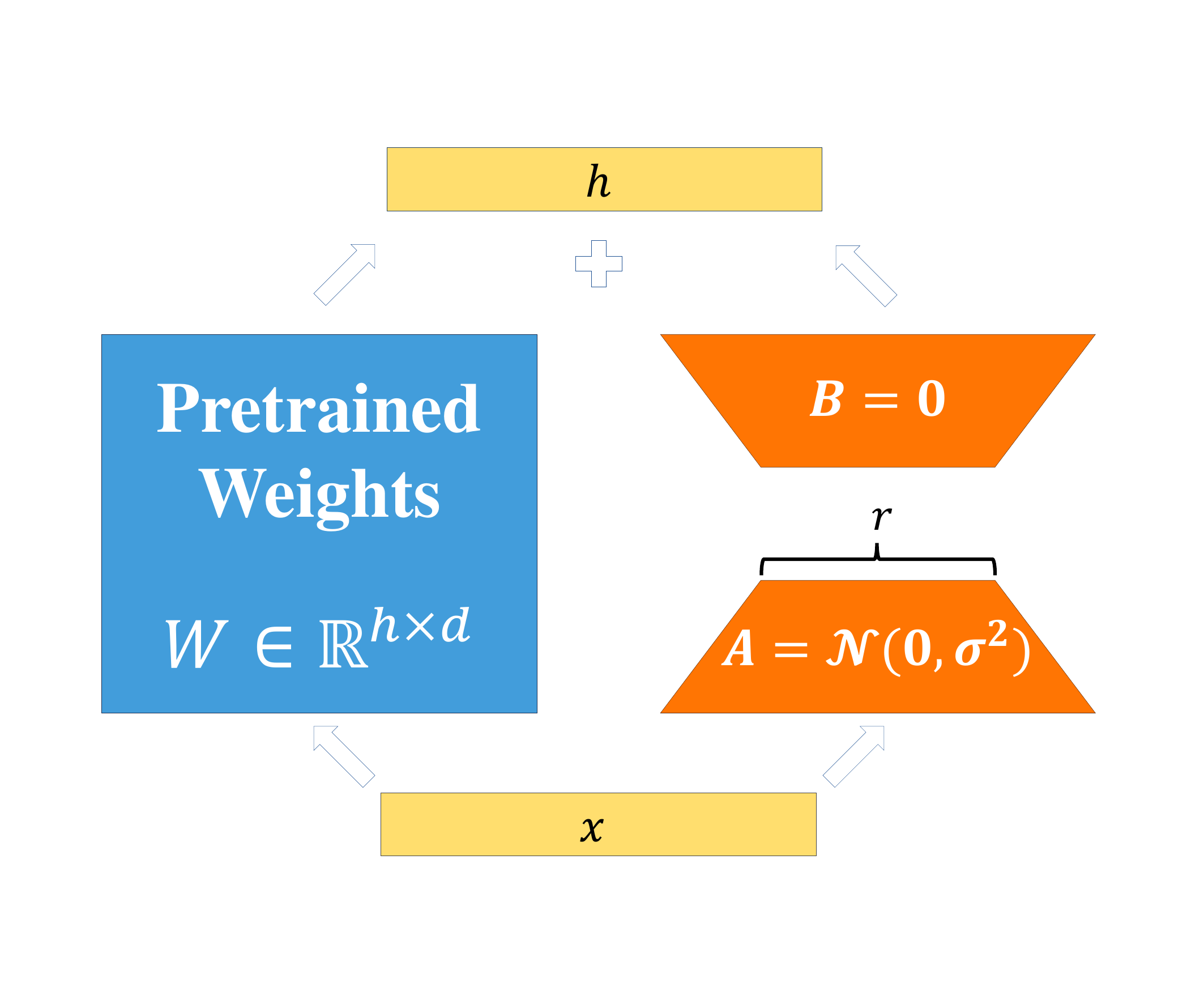} }}
    \subfloat[\centering BatchLoRA Approach]{{\includegraphics[width=.24\textwidth]{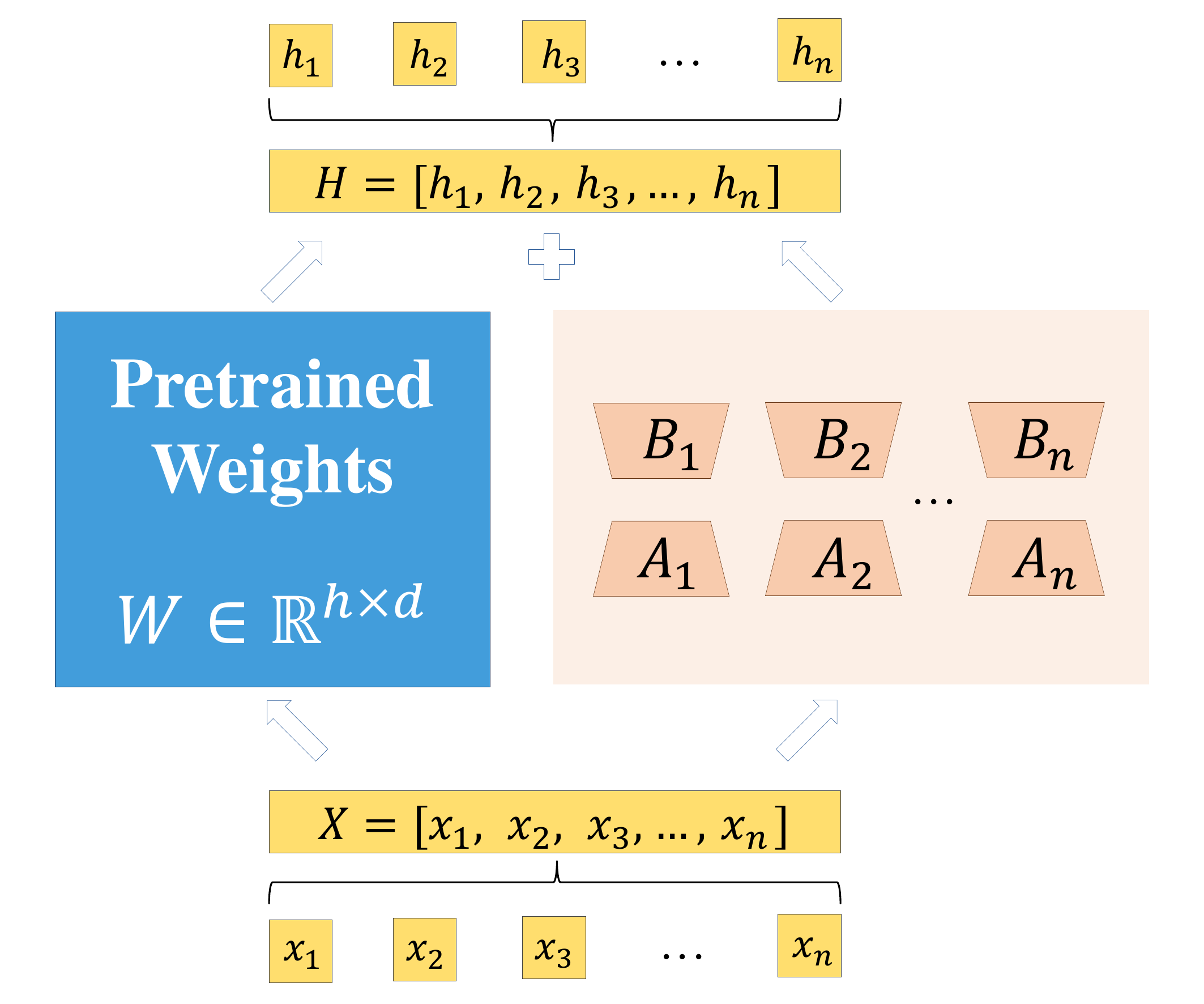} }}
    
    \caption{Sharing pre-trained model weights for fine-tuning multiple LoRA adapters with reduced overhead.}
    \label{fig:multi-lora}
\end{figure}

\section{Background and Motivation}
\label{sec:back}
\subsection{LoRA-based LLM Finetuning}
Training an LLM from scratch demands extensive computational resources over days of time, often utilizing thousands of GPUs and incurring significant financial costs~\cite{jetmoe,vtrain}. In contrast, fine-tuning pre-trained language models (PLMs) has made LLM benefits more accessible. Organizations like Meta and Google provide their PLMs, such as LLaMA ~\cite{llama} and Gemma~\cite{gemma}, to the public. Fine-tuning these models for various downstream tasks is effective~\cite{review-of-llm} and offers a more cost-efficient way to harness LLM capabilities.

Conventionally, full-weight fine-tuning of large-scale pre-trained models requires updating all parameters, which often incurs prohibitive computational costs. In contrast, Parameter-Efficient Fine-Tuning (PEFT) methods~\cite{peft} selectively update only a small subset of parameters, significantly reducing computational and memory resources. LoRA~\cite{lora}, a state-of-the-art PEFT technique, achieves efficient fine-tuning by freezing the pre-trained model and only updating low-rank additive matrices with far fewer parameters, as expressed in Equation \ref{eq:lora}.
\begin{equation}
\label{eq:lora}
h = xW^{'} = x(W + AB) = xW + xAB
\end{equation}

Where $x$ denotes the input data, $W \in \mathbb{R}^{h \times d}$ represents the frozen pre-trained model weights, and $A \in \mathbb{R}^{h \times r}$ and $B \in \mathbb{R}^{r \times d}$ are two low-rank decomposition matrices, with rank $r \ll \text{min}(h, d)$. 

Figure ~\ref{fig:multi-lora}(a) shows a typical way to train a single LoRA adapter from a frozen PLM. When training multiple LoRA adapters simultaneously, it makes intuitive sense to {\em share the same read-only base model} among them to reduce the GPU memory footprint, as shown in Figure ~\ref{fig:multi-lora}(b). 
A naive implementation for such simultaneous fine-tuning is listed in Algorithm ~\ref{lst:SimpleAlgo}: It keeps the base model on the GPU throughout the entire training process for all LoRA tasks, {\em only} swapping the adapter weights for each task sequentially.

\begin{lstlisting}[caption={Simply train multiple LoRAs, PyTorch-like.}, label={lst:SimpleAlgo}]
for adapter, data in fine_tuning_task:
    A, B = adapter # swap in the low-rank matrix A and B
    output = data @ W + data @ A @ B
    loss = loss_fn(data, output)
    loss.backward()
\end{lstlisting}

\begin{lstlisting}[caption={Use the BatchLoRA to train, PyTorch-like.}, label={lst:BatchLoRA}]
datas = [data for _, data in fine_tuning_task]
adapters = [adapter for adapter, _ in fine_tuning_task]
output = datas @ W # just call once
output += BatchLoRA.apply(datas, adapters)
loss = loss_fn(data, output)
loss.backward()
\end{lstlisting}

\subsection{Multi-LoRA Finetuning across Multi-GPU}
\label{sec:mm}
When the need to fine-tune multiple LoRA adapters exceeds the capacity of a single GPU -- mainly due to limited GPU memory and/or computation -- parallelization through multiple GPUs is necessary. Two common parallelism methods are {\em data parallelism} (DP)~\cite{data-parallel} and {\em model parallelism}~\cite{megatron-lm}. Data parallelism requires each GPU to store a complete set of model parameters, which is inefficient and even impossible for LLM training/fine-tuning when the model size is large and the GPU memory is small. For example, we cannot fine-tune a Llama-2-13B model in fp32 precision using FSDP~\cite{fsdp} with 4 $\times$ NVIDIA RTX A6000 48GB GPUs.

To address this, model parallelism partitions and distributes model parameters across GPUs. {\em Tensor parallelism} (TP), one of the representative model parallelism strategies, splits a tensor (e.g., a vector or matrix) in the model into multiple chunks along a specific dimension. Each GPU only holds one chunk of the tensor and computes partial results based on the allocated tensor chunk. All partial results are combined into the final result through collective communication methods, such as all-reduce or all-gather. However, this approach introduces significant synchronization overhead, particularly in inter-machine setups, where limited communication bandwidth can substantially slow down LLM training.

To mitigate this, pipeline parallelism (PP) divides the model into sequential groups, each containing one or more layers of the model. Each GPU handles a separate group and computationally depends on its previous GPU, which manages the preceding group. Consequently, input data is processed in a sequential, pipelined manner, passing through the dependent GPUs. PP reduces communication overhead by transmitting only the results of the last layer in a group between adjacent GPUs rather than synchronizing the intermediate results of each tensor within each layer. Nevertheless, GPU idle times can be significant due to the computational dependencies of PP. Solutions like PipeDream~\cite{pipedream} and PipeMare~\cite{pipemare} relax dependency constraints, e.g., using mismatched weight versions between forward and backward propagation, to reduce pipeline bubbles. However, recent works~\cite{pp-breaking,pp-asyn-decentralized,pp-swarmsgd} suggest that these methods may lead to lower convergence performance. In the context of LoRA-based fine-tuning, we have two key observations:

\begin{figure}[t]
    \centering
    \includegraphics[width=0.48\textwidth]{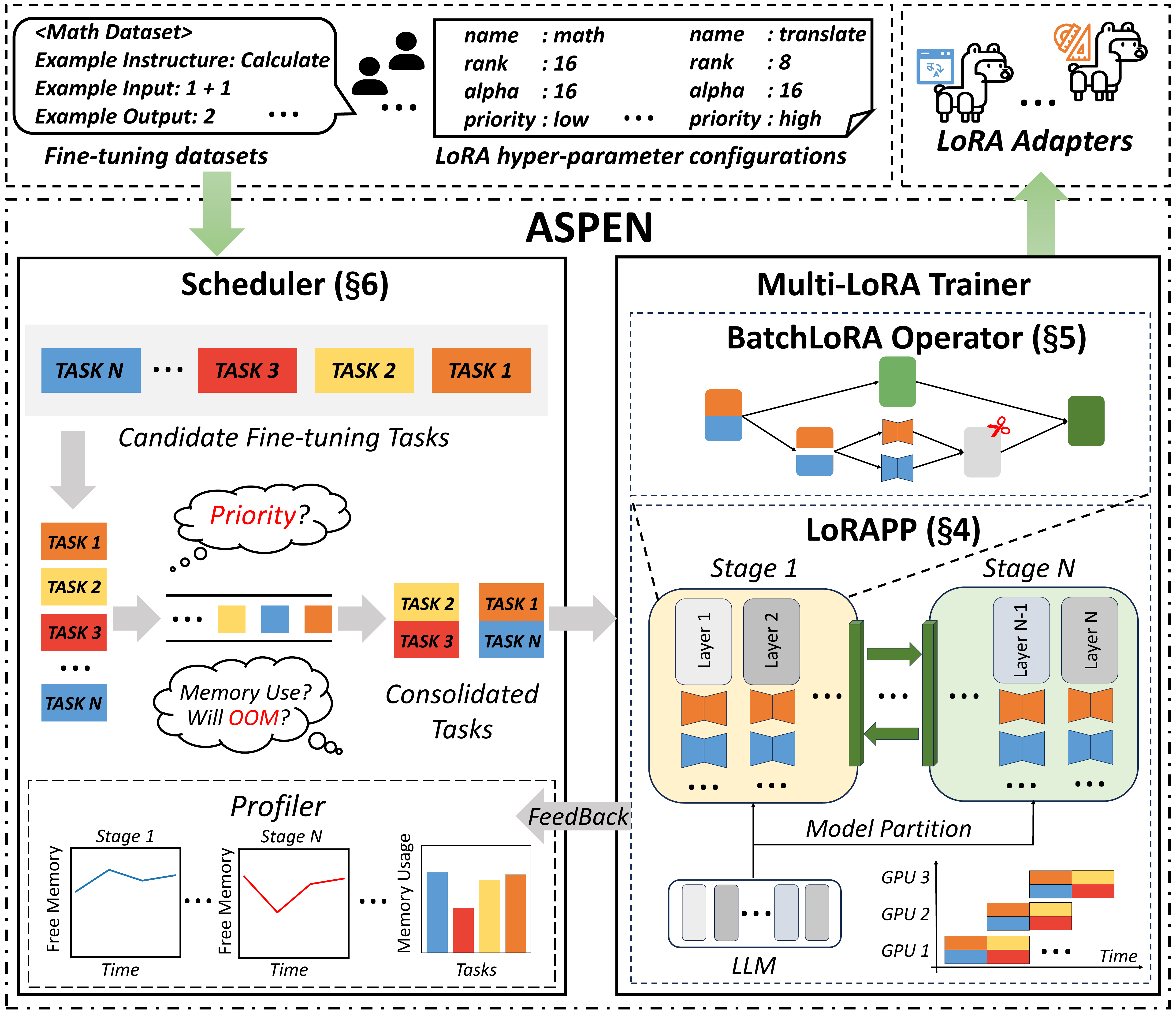}
    \caption{Overview of ~mLoRA.}
    \vspace{-0.15in}
    \label{fig:overview}
\end{figure}

{\em \underline{Observation 1:}} Unlike existing model parallelism strategies that require pipelining the {\em dependent} processing stages when training a single LLM, the {\em independent nature} of fine-tuning {\em multiple} LoRA adapters, despite sharing the same base model, can enable more efficient processing and greater parallelism. For example, we can populate a fully occupied fine-tuning pipeline across multiple GPUs and machines by scheduling distinct training stages for separate LoRA adapters {\em concurrently}. Further, by overlapping GPU communication and computation across separate stages, we can effectively hide I/O latencies and maximize overall GPU efficiency.

{\em \underline{Observation 2:}} The overhead from calling the CUDA API to launch GPU kernel functions can be substantial. This is particularly true when we fine-tune numerous small LoRA adapters with a naive parallel scheme like Algorithm ~\ref{lst:SimpleAlgo}, which leads to frequent kernel launches and high overhead, e.g., accounting for up to 10\% of the total training time. 
A promising solution to mitigate this overhead, as illustrated in Figure~\ref{fig:multi-lora}(b) and Algorithm ~\ref{lst:BatchLoRA}, is to consolidate the training data from multiple fine-tuning tasks into a larger batch. By performing matrix operations for all involved adapters collectively, we can achieve the same results as executing multiple fine-tuning tasks sequentially (as that in Algorithm ~\ref{lst:SimpleAlgo}) but with fewer GPU kernel launchers and reduced overall training time.

\section{Design of mLoRA}
The limitations of existing model parallelism methods and the observations in Section~\ref{sec:mm} motivate us to design mLoRA, a new fine-tuning system for the efficient training of multiple LoRA adapters. In this section, we first present an overview of mLoRA, including its key design objectives and fine-tuning workflow, and then detail the key techniques that underpin mLoRA.

\subsection{Overview}
\noindent \textbf{Design objectives:}
mLoRA is developed to fine-tune multiple LoRA adapters efficiently across one or multiple (cost-effective) GPUs. It optimizes training throughput and resource utilization via two new techniques: 1) LoRA-aware pipeline parallelism, LoRAPP (\S\ref{sec:lorapp}), and 2) LoRA-efficient training operator, BatchLoRA (\S\ref{sec:batchlora}). 

\smallskip
\noindent\textbf{Architecture Overview:}
As illustrated in Figure \ref{fig:overview}, mLoRA consists of two main components:
1) A \textit{multi-LoRA trainer} capable of simultaneously handling multiple LoRA fine-tuning tasks while conducting runtime optimization via BatchLoRA and LoRAPP technologies. 
2) A task scheduler that can choose a batch of fine-tuning tasks based on user demands and metrics from the performance profiler, e.g., to schedule tasks to maximize GPU resource utilization and minimize the out-of-memory (OOM) issues.

Specifically, users initiate requests to ~mLoRA, providing hyper-parameter configurations for the LoRA adapters and the datasets used for fine-tuning. Based on this, ~mLoRA generates candidate tasks with their initial configurations and places them in a candidate task queue.
Then, the task scheduler chooses tasks from the candidate task queue for parallel training (\S~\ref{sec:lorapp} and \S~\ref{sec:batchlora}) based on various scheduling factors (\S~\ref{sec:scheduler}), such as the memory footprint and task priority. During the training, the multi-LoRA trainer provides performance metrics to the profiler, including the actual memory usage of the current task. The profiler then uses this information to keep revising its memory estimation model (\S~\ref{sec:scheduler}), enabling more precise assessments of memory requirements for future tasks.

\subsection{Multi-LoRA Training Parallelism}
\label{sec:lorapp}

\subsubsection{LoRA-aware Pipeline Parallelism (LoRAPP)}
\begin{figure*}[htb]
    \centering
    \includegraphics[width=1.0\textwidth]{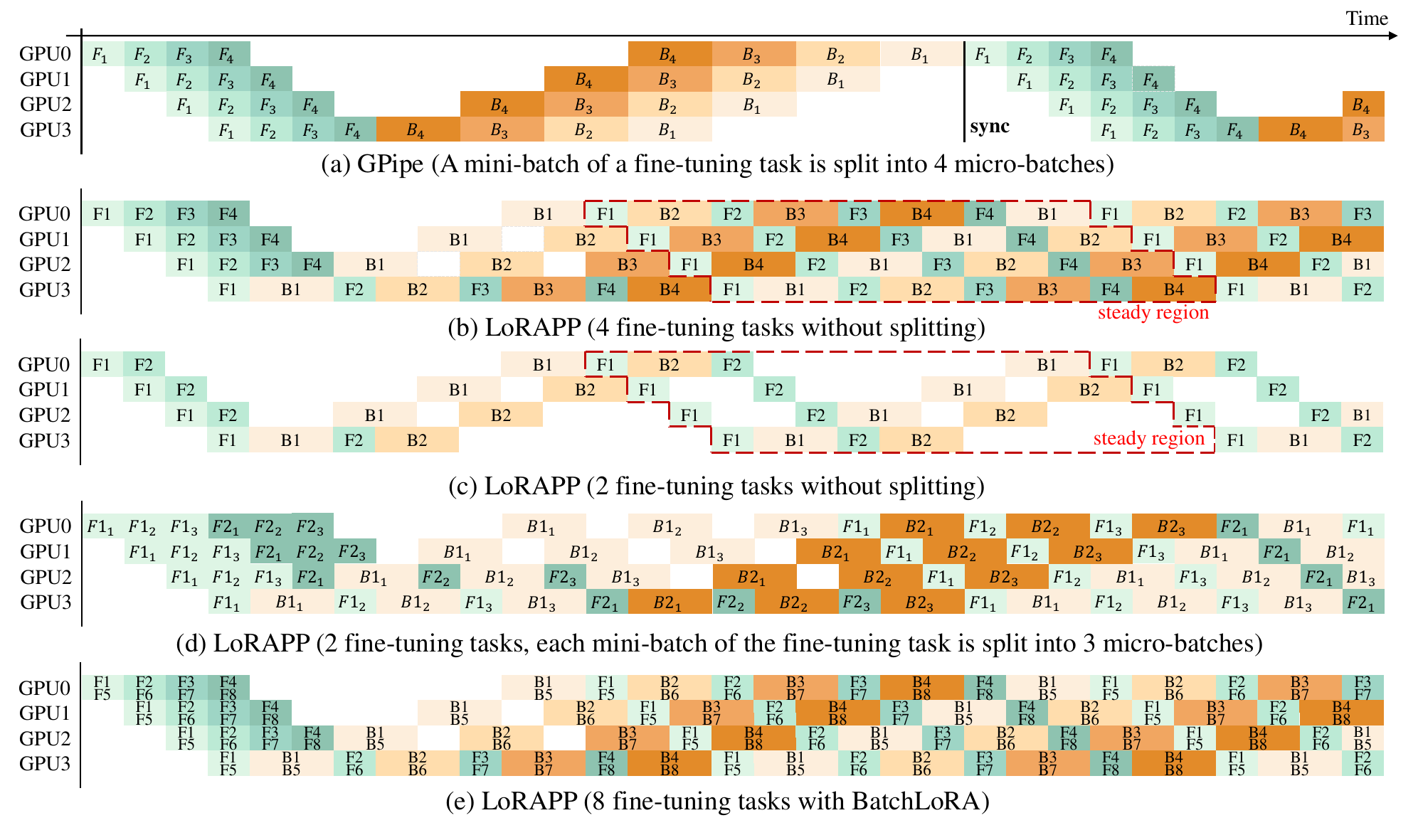}
    \caption{(a) GPipe. The training data of the fine-tuning task are divided into four micro-batches within a mini-batch. Here, $F_i$ represents the forward propagation of the $i$th micro-batch, while $B_i$ represents its backward propagation. GPipe requires all micro-batches of the same mini-batch to be completed before proceeding to the next mini-batch. (b) (c) LoRAPP without mini-batch splitting. $Fi$ represents the forward propagation of the $i$th LoRA adapter, while $Bi$ is its backward propagation. (d) LoRAPP with mini-batch splitting. $Fi_j$ represents the forward propagation of the $j$th mini-batch, into which the macro-batch of training data for the $i$th LoRA adapter is divided, while $Bi_j$ represents its backward propagation. (e) LoRAPP with BatchLoRA.}
    \label{fig:pipe}
\end{figure*}

As discussed in Section~\ref{sec:back}, pipeline parallelism can lead to idle periods and inefficiencies due to computational dependencies between GPUs. 
For example, in Figure ~\ref{fig:pipe} (a), the traditional pipeline parallel algorithm GPipe~\cite{gpipe} requires $GPU0$ to wait for $GPU1$ to complete $B1$ before $GPU0$ can execute $B1$, creating idle times for $GPU0$, known as {\em pipeline bubbles}. Drawing on Observation 1 (\S~\ref{sec:mm}), we propose LoRAPP, a novel pipeline parallelism strategy to optimize fine-tuning multiple LoRA tasks by reducing or eliminating these bubbles.

\begin{figure}[t]
    \centering
    \includegraphics[width=0.48\textwidth]{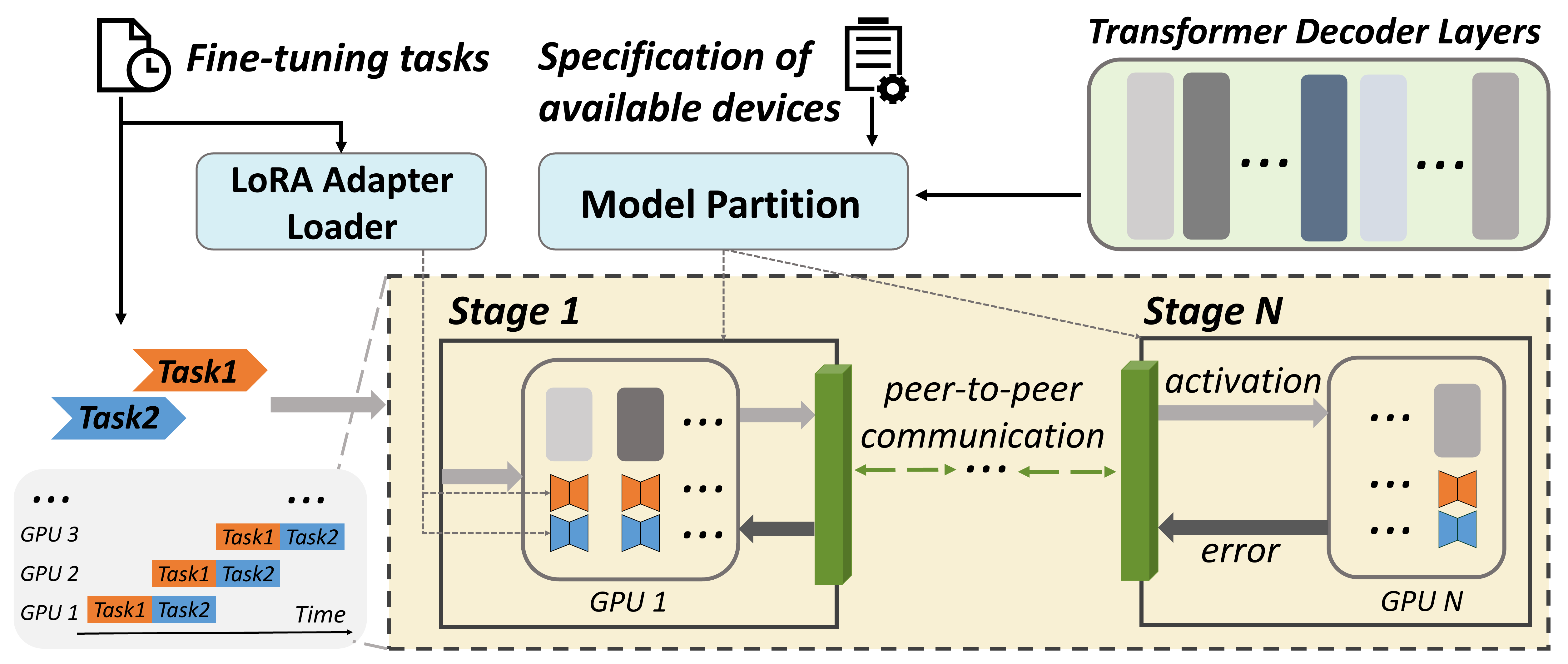}
    \caption{The workflow of LoRAPP.}
    \label{fig:pipe-workflow}
\end{figure}

\smallskip
\noindent\textbf{Base workflow of LoRAPP.} 
As illustrated in Figure ~\ref{fig:pipe-workflow}, the workflow of LoRAPP comprises two main stages. 

In the {\em preparation} stage, LoRAPP partitions the pre-trained base LLM -- comprising consecutive transformer decoder layers -- into separate groups and allocates these groups to available GPUs (e.g., one group for each GPU). Note that model partitioning is not the focus of mLoRA and has been extensively covered in recent work~\cite{fan2021dapple, pipedream,gpipe}; LoRAPP adopts the partitioning approach from GPipe to ensure that each group has an equal computational load. 

In the {\em training} stage, following mLoRA's scheduling scheme (\S ~\ref{sec:scheduler}), a set of fine-tuning tasks is selected for parallel training and populating the multi-GPU pipeline: 1) For each LoRA adapter, each GPU allocates a small amount memory to store a portion of the adapter's weights associated with the linear layers of the base model assigned to the current GPU. These weights are randomly initialized as described in LoRA~\cite{lora}. 2) After initialization, each GPU performs {\em forward propagation} using activation values received from its previous GPU's forward propagation. 3) After forward propagation, each GPU performs {\em backward propagation} using error values received from its next GPU's backward propagation. 

During the pipelined processing, the first and last GPUs operate slightly differently from others: 1) The first GPU in the pipeline receives the training data for a fine-tuning task to initiate the training process and does not need to send error values. 2) The last GPU computes the loss using the activation values and then begins the backward propagation, without needing to send activation values. Once the fine-tuning task is finished, the weights of its LoRA adapters are saved (e.g., to persistent storage), and the allocated memory spaces can be released and used for new tasks. 
\smallskip 

\noindent{\bf Achieving Zero Bubbles in LoRAPP.}
\label{sec:bubble}
A key goal of LoRAPP is to reduce or eliminate pipeline bubbles and achieve high efficiency in pipelined fine-tuning.
Existing pipeline approaches, like GPipe~\cite{gpipe} as illustrated in Figure ~\ref{fig:pipe} (a), address this by dividing a mini-batch into smaller micro-batches to populate the pipeline during each training step or iteration. However, to ensure model convergence, the mini-batch gradient descent algorithm~\cite{mini-batch-sgd} requires that the pipeline waits for gradients from all micro-batches within a mini-batch to accumulate before applying them. This {\em stop-and-wait} synchronization introduces pipeline bubbles. While increasing the number of micro-batches can alleviate the pipeline bubbles to some extent, the micro-batch count is constrained by the mini-batch size. Moreover, larger mini-batch sizes can negatively impact model convergence ~\cite{ben2019demystifying,cheng2021dataset}, further restricting the mini-batch size. As a result, it is hard for existing pipeline parallel approaches to achieve zero pipeline bubbles while ensuring model convergence.

In contrast, LoRAPP reduces the pipeline bubble to {\em zero} based on Observation 1 (\S~\ref{sec:mm}): Since each LoRA adapter independently accumulates and applies gradients, there is no need to synchronize gradients between different LoRA adapters. Thus, LoRAPP can use mini-batches from different LoRA adapters to populate the pipeline. For example, in Figure ~\ref{fig:pipe} (b), after $GPU0$ completes the forward propagation $F1$ of LoRA adapter 1, it immediately begins the forward propagation $F2$ of LoRA adapter 2. 
When $GPU0$ completes the backward propagation $B1$ of LoRA adapter 1, it can immediately perform the forward propagation $F1$ using the next mini-batch data of LoRA adapter 1. As shown in Figure ~\ref{fig:pipe} (b), during the steady state, the pipeline is fully utilized by different forward/backward propagation processing of distinct LoRA adapters.
It is important to note that when reaching the zero bubble state as shown in Figure ~\ref{fig:pipe} (b), a GPU, such as $GPU3$ after completing $F1$, needs to choose between executing $F2$ and $B1$. Since backward propagation can release a significant amount of memory for activations, optimizations, and weight gradients, we prioritize executing backward propagation to free up memory to accommodate more fine-tuning tasks. 

One problem remains: LoRAPP cannot achieve zero bubbles with fewer fine-tuning tasks, as shown in Figure ~\ref{fig:pipe} (c). To overcome this, as illustrated in Figure ~\ref{fig:pipe} (d), within the same LoRA adapter, LoRAPP adopts the same strategy as GPipe, which divides the mini-batch into multiple (e.g., three) micro-batches to reduce the bubbles.

The independence of training multiple LoRA adapters also enables the opportunity to overlap GPU communication and computation. As illustrated in Figure ~\ref{fig:communication-overlapping}, since there is no dependency between the $i$th and $j$th LoRA adapters, while the $i$th LoRA adapter's backward propagation $B_i$ is being executed on GPU $K+1$, it can simultaneously receive the $j$th LoRA adapter's forward propagation $Fj$ from GPU $k$. Such overlapping can greatly hide the I/O latency from GPU computation, further improving the efficiency of LoRAPP. More concretely, we create three independent and concurrent running CUDA streams for each GPU, each dedicated to receiving, sending, and computing data.

\subsubsection{Cost Analysis of LoRAPP}
\label{sec:lora_pp_cost}
To quantify the overhead introduced by pipeline bubbles in LoRAPP, we define the \emph{bubble ratio} as the ratio of GPU \emph{idle time} to the total \emph{runtime} of the pipeline.
\smallskip

\noindent\textbf{Bubble ratio in LoRAPP.} As shown in Figure ~\ref{fig:pipe} (c), each subsequent region repeats the steady region, so we can measure the bubble ratio by focusing on one steady region.
We define the forward propagation time as $T_f$, and the backward propagation time as $T_b$, with a total of $D$ GPUs training $L$ tasks simultaneously. Then, the total time of the steady region is $D^2(T_f+T_b)$, and the idle region is $max\{D(T_f+T_b)(D-L), 0\}$. 

Therefore, the bubble ratio of LoRAPP without using mini-batch splitting is $max\{(D-L)/D, 0\}$. 
Similarly, we can obtain the bubble ratio of GPipe as $(D-1)/(N+D-1)$, where $N$ represents the number of micro-batches.
This means that if the number of LoRA adapters trained in parallel is greater than or equal to the number of GPUs, LoRAPP can fill the pipeline to fully utilize all GPUs. As mentioned earlier, the number of macro-batches $N$ usually has a small value, preventing GPipe from achieving a zero bubble ratio.

When the system has more GPUs and fewer LoRA adapters for fine-tuning, LoRAPP achieves a relatively high bubble ratio. To further decrease the bubble ratio, as shown in Figure ~\ref{fig:pipe} (d), LoRAPP adopts the same strategy as GPipe. This way, its bubble ratio is $max\{(D-1+N-L \times N)/(D+N-1), 0\}$.
\smallskip

\begin{figure}[t]
    \centering
    \includegraphics[width=0.475\textwidth]{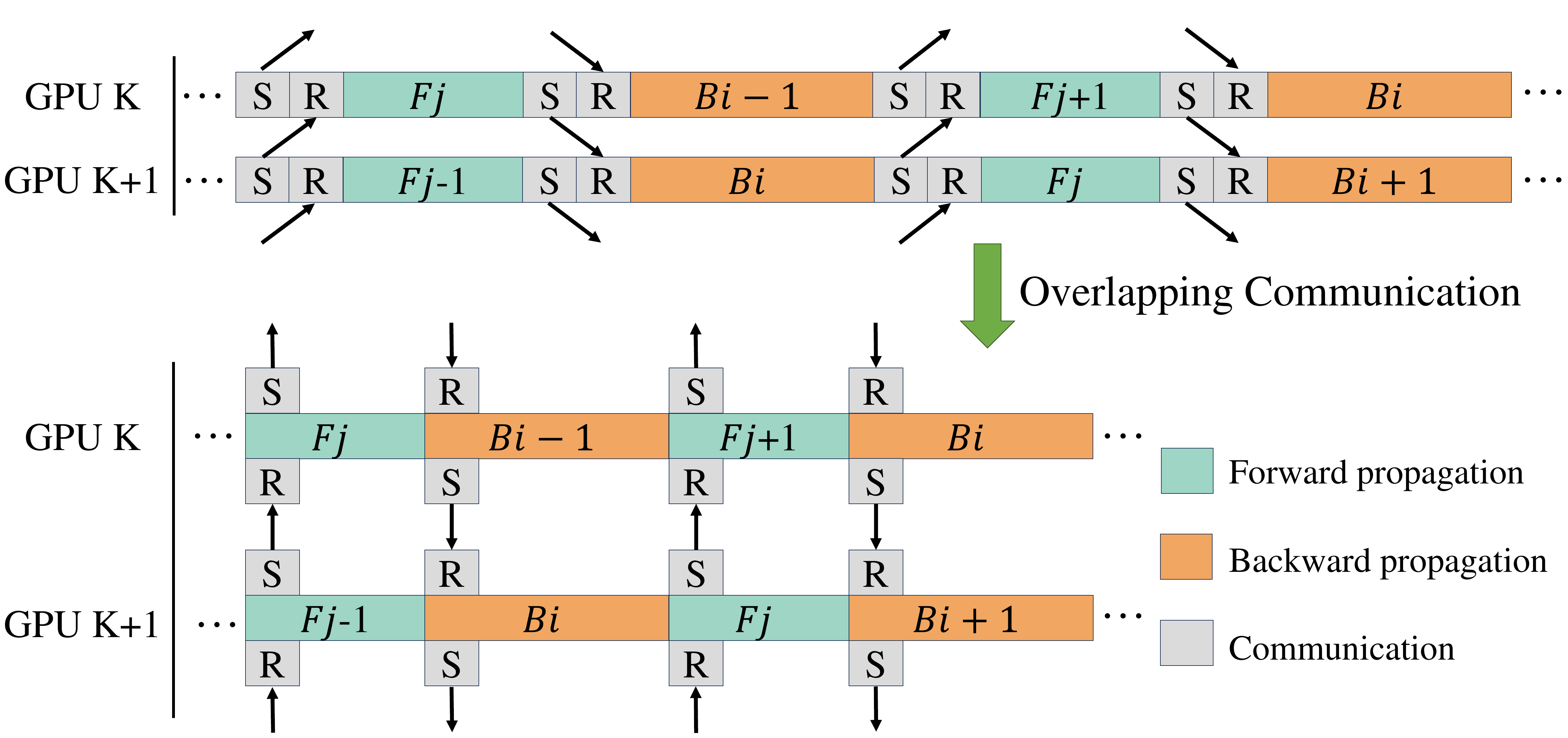}
    \caption{Overlapping communication in LoRAPP. $Fi$ represents the forward propagation of the $i$th LoRA adapter, while $Bi$ is its backward propagation.}
    \label{fig:communication-overlapping}
\end{figure}

\noindent\textbf{Communication cost.} In LoRAPP, data communication occurs when activation and error values are exchanged between partitions.
Therefore, the communication volume depends on the number of partitions, which is the number of GPUs $D$, and the size of the input data. We use $B$ to represent the number of input data tokens and $h$ to represent the model's hidden size. The size of activation values and error values is denoted using $Bh$. Hence, the total communication volume is $2(D-1)Bh$. 
As shown in Figure ~\ref{fig:communication-overlapping}, mLoRA overlaps communication and computation to hide such communication latency overhead.
\smallskip

\noindent\textbf{Performance analysis.} 
As mentioned before, LoRAPP does not incur additional computational overhead compared to GPipe and its communication can be overlapped. Thus, its throughput can be roughly estimated as $R{(1 - \nu)}/{(1 - \mu)}$, where $R$ is the throughput of GPipe, $\mu$ is the bubble ratio of GPipe, and $\nu$ is bubble ratio of LoRAPP. Therefore, the throughput of LoRAPP is $R \times L$ when the zero bubble state is not reached, otherwise $R(D+N-1)/N$.
\smallskip

\noindent\textbf{Memory usage.} One key difference between LoRAPP and GPipe when training $L$ number of LoRA adapters is that LoRAPP shares the base model among these adapters. Therefore, LoRAPP saves $(L-1)W_\theta$ memory, where $W_\theta$ is the size of the pre-trained model.

\subsection{Multi-LoRA Training Operator}
\label{sec:batchlora}
With zero bubbles and hiding I/O communication latency, LoRAPP (\S~\ref{sec:lorapp}) achieves efficient pipelined fine-tuning across multiple GPUs. However, we observe that the pipelined GPUs remain not fully utilized. One reason lies in that, unlike complex tasks (e.g., full-fledged LLM training), each LoRA fine-tuning task (i.e., forward or backward propagation) performed by a GPU is relatively simple and cannot fully exploit the parallel processing capabilities of the GPU. As shown in Figure ~\ref{fig:pipe} (b), though theoretically, {\em four} fine-tuning tasks can reduce the pipeline's bubble to zero with {\em four} GPUs (i.e., according to the bubble ratio in \S~\ref{sec:lora_pp_cost}), a single GPU only uses part of its computation resources practically. For example, with the workload and single-machine multi-GPU setup in Section ~\ref{sec:exp_setup}, using Llama-2-7B as the base model with four fine-tuning tasks, the average GPU utilization is 83\%, and the average memory utilization is only 30\%.

To further improve GPU efficiency and utilization, one intuitive approach is to maximize the number of distinct fine-tuning tasks in the LoRAPP pipeline by scheduling as many LoRA adapters as possible. Note that the maximum number of LoRA adapters each GPU can handle is constrained by its memory size. However, as highlighted in Observation 2 from Section~\ref{sec:mm}, the overhead from calling CUDA APIs to launch  GPU kernel functions can be nontrivial when training numerous small LoRA adapters (with Algorithm~\ref{lst:SimpleAlgo}). To address this, mLoRA introduces a new operator, BatchLoRA, which allows multiple LoRA adapters to {\em concurrently} share the pre-trained base model with reduced kernel launch overhead.

\subsubsection{BatchLoRA Operator}
\label{sec:operator}

As illustrated in Algorithm~\ref{lst:BatchLoRA} and Figure~\ref{fig:multi-lora}(b), \emph{BatchLoRA} consolidates the training data for a selected number of LoRA fine-tuning tasks into a single large batch (i.e., a large matrix) during each training iteration. Therefore, multiple LoRA adapters can share the same pre-trained model and participate in training {\em concurrently} -- instead of sequentially like Algorithm~\ref{lst:SimpleAlgo}. 

We use Figure ~\ref{fig:multi-lora} (b) as the running example. Suppose a set of fine-tuning tasks, denoted as $T_1, ..., T_n$. Each fine-tuning task, $T_i$, consists of the fine-tuning input data represented as $x_i$, along with the low-rank weights $A_i$ and $B_i$ of the LoRA adapters. Note that, all the fine-tuning tasks share the same pre-trained weights $W$. Formally, given the input data $x_i$ for the i-th fine-tuning task and the output data $h_i$, the consolidated input data $X=({x_1}^\intercal,\dots,{x_n}^\intercal)^\intercal$. The calculation formula for forward propagation is shown as Formula~\ref{eq:forward}.

\begin{equation}
\label{eq:forward}
H = \begin{pmatrix} h_1 \\ \vdots \\ h_n \end{pmatrix} = \begin{pmatrix}
    x_1 \\ \vdots \\ x_n
\end{pmatrix}W + \begin{pmatrix}
    x_1A_1B_1 \\
    \vdots \\
    x_nA_nB_n
\end{pmatrix} = XW + \begin{pmatrix}
    x_1A_1B_1 \\
    \vdots \\
    x_nA_nB_n
\end{pmatrix}
\end{equation}

For backward propagation, according to Formula ~\ref{eq:forward}, we derive the gradient formula for each tensor involved in the computation as Formula~\ref{eq:grad_lora} and \ref{eq:grad_output}. Note that $W$, i.e., the frozen pre-trained weights, does not require training, so its gradients do not need to be computed.

\begin{equation}
\label{eq:grad_lora}
\begin{pmatrix} \nabla A_1 \\ \vdots \\ \nabla A_n \end{pmatrix} = \begin{pmatrix}
    {x_1}^\intercal \nabla h_1{B_1}^\intercal \\
    \vdots \\
    {x_n}^\intercal \nabla h_n{B_n}^\intercal 
\end{pmatrix} 
\ \ , \ \ 
\begin{pmatrix}
    \nabla {B_1} \\
    \vdots \\
    \nabla {B_n}
\end{pmatrix} = \begin{pmatrix}
    {A_1}^\intercal{x_1}^\intercal \nabla h_1 \\
    \vdots \\
    {A_n}^\intercal{x_n}^\intercal \nabla h_n
\end{pmatrix}
\end{equation}

\begin{equation}
\label{eq:grad_output}
\nabla X = \begin{pmatrix}
    \nabla x_1 \\
    \vdots \\
    \nabla x_n
\end{pmatrix} =
\nabla H W^\intercal + \begin{pmatrix}
    \nabla h_1 {B_1}^\intercal {A_1}^\intercal \\
    \vdots \\
    \nabla h_n {B_n}^\intercal {A_n}^\intercal
\end{pmatrix},
\nabla H = \begin{pmatrix}
    \nabla h_1 \\
    \vdots \\
    \nabla h_n
\end{pmatrix}
\end{equation}

Therefore, based on Formula ~\ref{eq:forward} and ~\ref{eq:grad_output}, we can find that after the training data is consolidated, we only need to launch the matrix multiplication operation $XW$ and $\nabla H W^\intercal$ once on the GPU, rather than launching the matrix multiplication operation ${x_i}W$ and $\nabla h_i W^\intercal$ for each LoRA adapter, thereby reducing the overhead of kernel launches. Note that training with consolidated data does not affect the model performance and isolation between different fine-tuning tasks, since each LoRA adapter only uses the specific portion of the training data that belongs to this adapter for computation.
\smallskip 

\begin{figure}[t]
    \centering
    \includegraphics[width=0.48\textwidth]{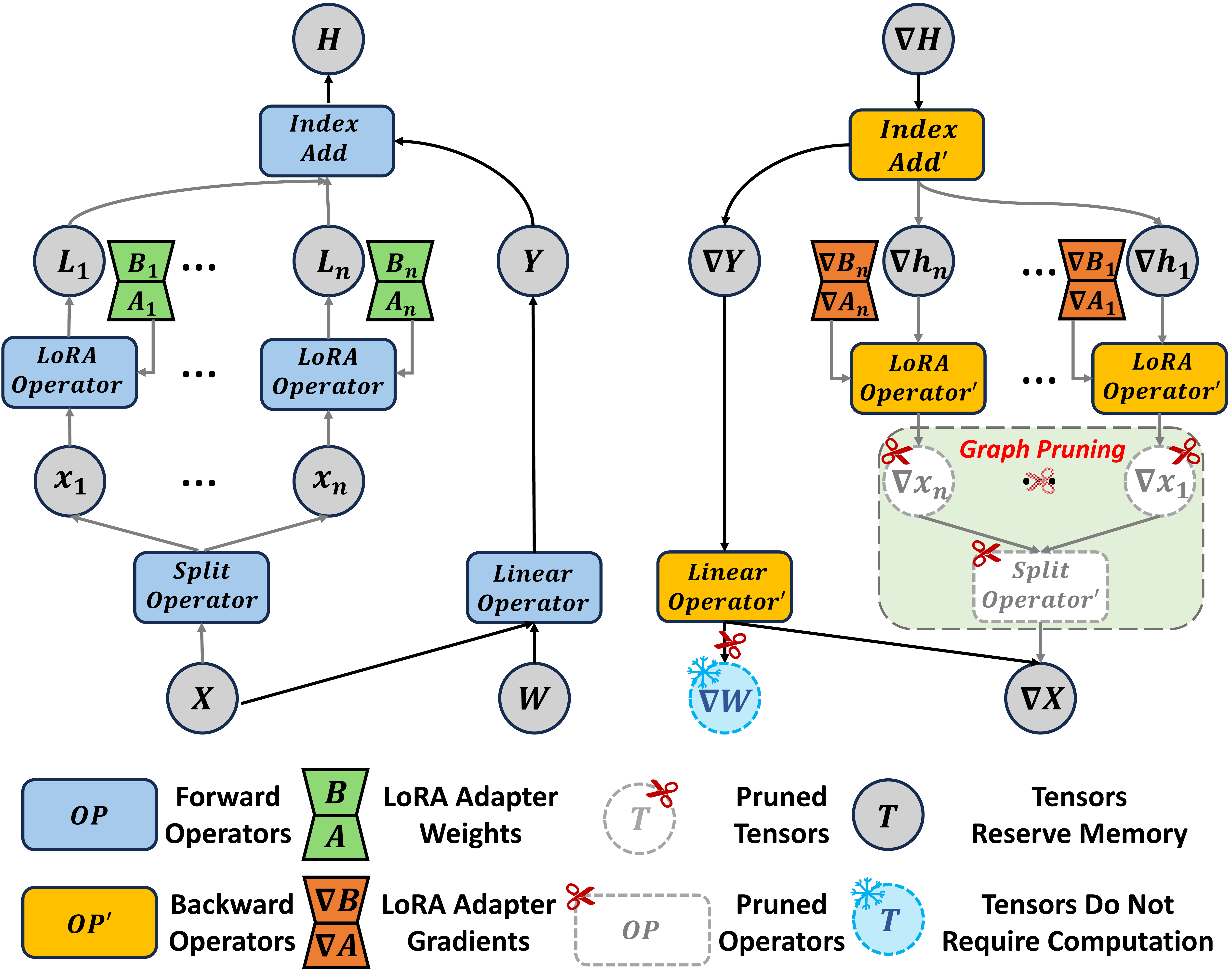}
    \caption{Computational graphs of BatchLoRA operator with graph pruning.}
    \label{fig:compute_graph}
\end{figure}

\noindent{\bf Workflow of BatchLoRA.}\label{sec:graph_pruning}
\label{sec:graphpruning}
mLoRA follows existing reverse-mode automatic differentiation and gradient computation, implemented through computational graphs~\cite{pytorch}. It automatically determines a backward propagation computational graph based on the forward propagation computational graph (defined by the user) and then computes the gradients through this graph. 
For example, the computational graph of the BatchLoRA operator, as shown in Figure ~\ref{fig:compute_graph}, consists of two parts: the forward propagation defined by the user (i.e., the left diagram) and the backward propagation automatically determined (i.e., the right diagram).

For BatchLoRA's forward propagation, the consolidated input data $X$ is used to compute intermediate results $Y=XW$ with the frozen pre-trained weights $W$. Then, since the consolidated data $X$ represents the training data for multiple LoRA adapters, we need to split it into multiple chunks ${x_1,\dots,x_n}$ to make sure that each chunk represents the training data for its corresponding LoRA adapter, i.e., ensuring isolation among tasks (and their users). These data are separately computed with their respective LoRA adapters, resulting in intermediate results $L_i=x_iA_iB_i$. Finally, the intermediate results $L_i$ are added to $Y$ based on their original positions before splitting to obtain the final output $H$.

The backward propagation of BatchLoRA consists of two parts. In the first part, the gradients of the LoRA adapter $\nabla A_i$ and $\nabla B_i$ are computed as follows: Based on the split-position information during the forward propagation, the input backward propagated error $\nabla H$ is split into multiple chunks ${\nabla h_1,\dots, \nabla h_n}$, where each chunk represents the input backward propagated error for each LoRA adapter. Then, according to Formula ~\ref{eq:grad_lora}, the gradients for each LoRA adapter are computed separately.
In the second part, the output backward propagated error values $\nabla X$ are computed. According to Formula ~\ref{eq:grad_output}, first, the intermediate value $\nabla Y =\nabla H W^\intercal$ is computed; then the backward propagated error for each LoRA adapter is computed as $\nabla x_i = \nabla h_i {B_i}^\intercal {A_i}^\intercal$. Finally, the backward propagated errors ${\nabla x_1,\dots,\nabla x_n}$ are consolidated into an intermediate value through the derivative of the split operator and added to $\nabla Y$ to generate the final output backward propagated error $\nabla X$.

\smallskip 
\noindent\textbf{Graph pruning.} The backward propagation process, determined by the forward propagation computational graph, is usually suboptimal. mLoRA addresses this by constructing more efficient computational graphs to reduce unnecessary overhead rather than relying on the automatically generated computational graph. For example, Figure~\ref{fig:compute_graph}’s right diagram illustrates how mLoRA prunes the derivatives of the split operator within the backward propagation graph.
Once the intermediate values $\nabla Y$ and all backward propagated errors $\nabla x_i$ with LoRA adapters are computed, ~mLoRA adds $\nabla x_i$ to the corresponding positions of $\nabla Y$ using their positional information from the forward propagation, thus generating the final result $\nabla X$ and avoiding expensive memory operation overhead associated with split operator derivatives.
\smallskip 

\noindent\textbf{BatchLoRA-enahnced LoRAPP.} BatchLoRA complements LoRAPP to deliver highly efficient pipeline parallelism. As illustrated in Figure~\ref{fig:pipe} (e), mLoRA first aims for ``zero bubbles'' by matching the number of fine-tuning tasks to the number of GPUs whenever possible. BatchLoRA then consolidates any additional tasks to maintain this zero-bubble condition, ensuring that the number of combined tasks equals the number of GPUs. As discussed in Section~\ref{sec:scheduler}, mLoRA schedules as many tasks as the GPU memory allows, optimizing resource utilization.

\subsubsection{Cost Analysis}
\label{sec:analysis2}
To understand how BatchLoRA reduces overall training time for multiple fine-tuning tasks, we analyze its impact on minimizing the overhead associated with launching GPU kernel functions and the operational overhead introduced by BatchLoRA.
\smallskip

\noindent\textbf{Kernel launch cost.} As the cost of launching GPU kernel functions is proportional to the number of times the CUDA API is called~\cite{understanding-kernel-launch}, we define the kernel launch cost as the number of these calls. We assume that when fine-tuning one LoRA and conducting one complete forward and backward propagation, the kernel launch cost incurred by the pre-trained model's participation is $\alpha$, and the kernel launch cost for each LoRA adapter is $\beta$.

When fine-tuning $k$ LoRA adapters without using the BatchLoRA operator (Algorithm ~\ref{lst:SimpleAlgo}), each LoRA adapter and the pre-trained model conducts one complete forward and backward propagation using training data, resulting in the kernel launch cost of $k\alpha + k\beta$. When using the BatchLoRA operator (Algorithm ~\ref{lst:BatchLoRA}), the pre-trained model conducts one complete forward and backward propagation using the consolidated data, and each LoRA adapter conducts one complete forward and backward propagation using the training data, resulting in the kernel launch cost of $\alpha + k\beta$.

Therefore, BatchLoRA can reduce the kernel launch cost by approximately $((k-1)\alpha)/(k(\alpha + \beta))$. Since LoRA adapters hold significantly fewer parameters and matrix operations compared to the pre-trained model, it results in a much smaller cost, i.e., $\beta \ll \alpha$. Thus, the reduction in kernel launch cost is approximately $(k-1)/k$, where $k$ is the number of concurrently trained LoRA adapters.
\smallskip

\noindent\textbf{BatchLoRA operator cost.} The split operation pruned by the BatchLoRA operator does not alter the computational workload but reduces peak memory usage during the consolidation of multiple LoRA adapters. The memory savings equal the size of the input training data gradients, which matches the size of the input data. Assuming the total length of input tokens is $O$, and the hidden size of the model is $h$, it can save peak memory of $4Oh$ bytes in fp32 training precision. Moreover, it also reduces the latency associated with allocating and copying the redundant memory on GPUs.

\subsection{Task Scheduler}
\label{sec:scheduler}
The scheduling objective of mLoRA is to schedule as many fine-tuning tasks as possible for high system efficiency while satisfying user priorities and avoiding out-of-memory (OOM) errors~\footnote{Other scheduling strategies can be easily integrated into ~mLoRA.}. In this section, we first introduce mLoRA's preemptive priority scheduling to ensure user priorities and then describe how it avoids OOM and selects as many tasks as possible for concurrent execution.
\smallskip 

\noindent\textbf{Preemptive priority scheduling.} 
mLoRA uses a priority scheduling algorithm to address users' priority needs -- a common practice in multi-tenant environments. Each fine-tuning task is assigned a static priority, with the highest-priority tasks processed first. Tasks with the same priority are handled on a first-come, first-served basis. Scheduling decisions are made at the end of each iteration to promptly accommodate the preemption of high-priority tasks.
\smallskip 

\noindent\textbf{Modeling memory usage.} 
To achieve high parallelism and GPU efficiency, mLoRA schedules as many fine-tuning tasks as possible to maximize GPU memory utilization meanwhile avoiding OOM errors. To this end, mLoRA estimates the memory requirements of each fine-tuning task during task runtime. Specifically, mLoRA infers the relationship between memory size and the size of input training data as described in Vijay et al.~\cite{memory-estimate}. It conducts online model fitting in the following manner: 

\begin{equation}
\label{equ:online-function}
Mem = \beta_0 + \beta_1B_tL_n + \beta_2B_t{L_n}^2
\end{equation}

Where $Mem$ represents the required memory;  $L_n$ is the input training data sequence length; $B_t$ is the input batch size;  $\beta_0$, $\beta_1$, and $\beta_2$ are non-negative coefficients. Throughout the model training process, ~mLoRA continuously gathers data points ($B_t$, $L_n$, $Mem$) via the profiler (Figure~\ref{fig:overview}) and utilizes a non-linear least squares solver to determine the optimal coefficients for fitting this model~\cite{least_squares}.
In a single GPU setup, ~mLoRA only needs to ensure that the total memory required by all the scheduled fine-tuning tasks is less than the available memory to avoid OOM. 
In a multi-GPU setup with LoRAPP, ~mLoRA uses the model to estimate the required memory on each GPU and ensures that the estimated memory usage for each GPU is less than its available memory.

\section{Evaluation}
To demonstrate the effectiveness of mLoRA, we first evaluate the end-to-end performance in both single-GPU and multi-GPU environments with one or multiple machines (\S~\ref{sec:exp_real}). We then examine the benefits of the LoRAPP parallelism strategy (\S~\ref{sec:exp_lorapp}) and the BatchLoRA operator (\S~\ref{sec:exp_batchlora}), respectively.

\begin{figure*}[t]
    \centering
    \includegraphics[width=1.0\textwidth]{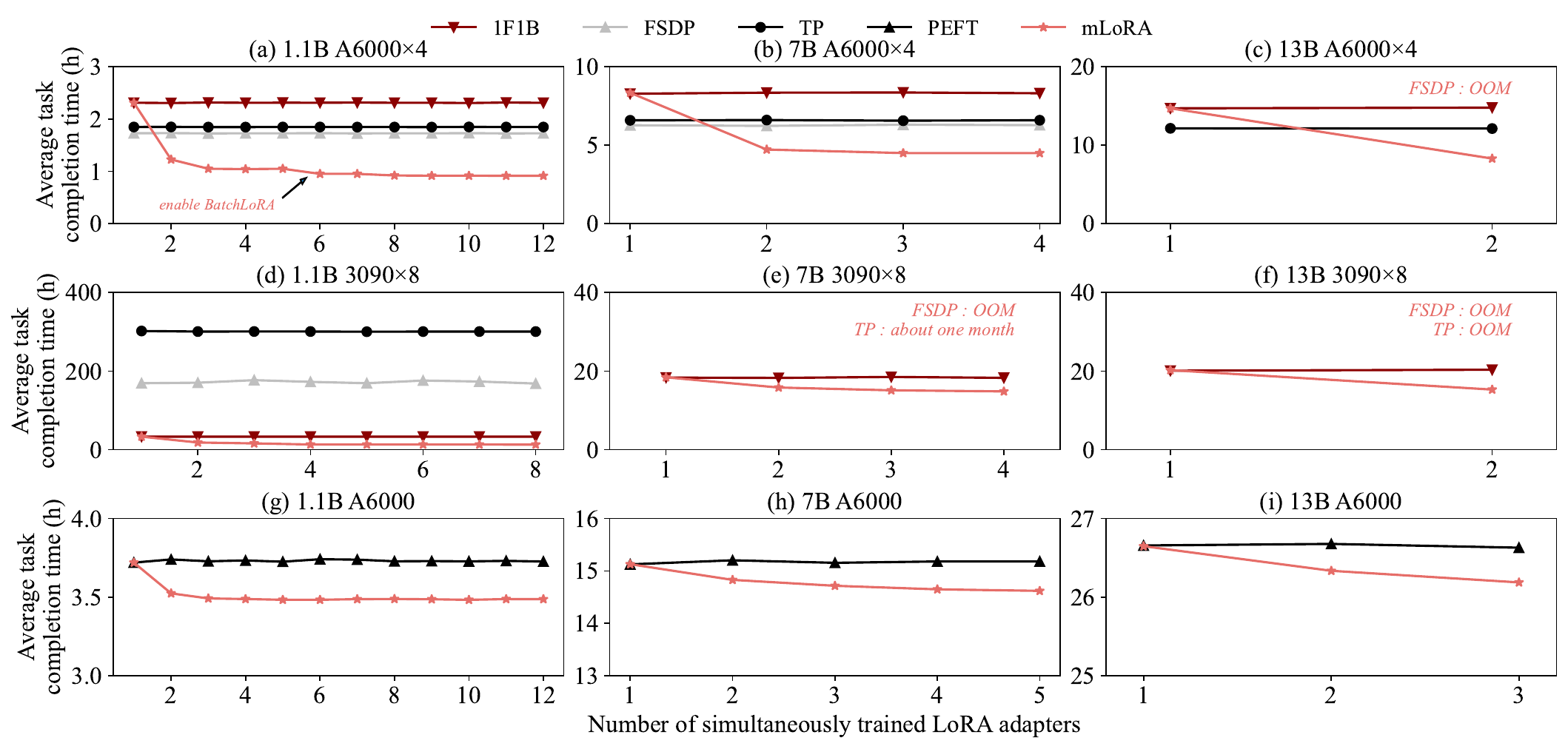}
    \caption{(a), (b), and (c) show the average fine-tuning task completion time in the single-machine multi-GPU setup. (d), (e), and (f) show the average fine-tuning task completion time in the multi-machine multi-GPU. (g), (h), and (i) show the average fine-tuning task completion time in the single-GPU setup. In the single-GPU setup, we can only run ~mLoRA using BatchLoRA; In the multi-GPU setup, we run ~mLoRA using BatchLoRA and LoRAPP. Note that, we enable BatachLoRA only when the number of fine-tuning tasks surpasses the number of GPUs to ensure zero pipeline bubbles.}
    \label{fig:end-to-end}
\end{figure*}

\subsection{Experimental Setup}
\label{sec:exp_setup}

\noindent\textbf{Models.} We evaluate ~mLoRA using three publicly accessible LLaMA model series, each with different parameter scales: Llama2-13B~\cite{llama2}, Llama2-7B, and TinyLlama-1.1B~\cite{tinyllama}.
\smallskip

\noindent\textbf{Platforms.} Our experimental platforms include both single-machine and multi-machine setups. In the single-machine setup, we use four (or eight) NVIDIA RTX A6000 GPUs, each with 48GB of memory, connected via PCIe 4.0x16. For the multi-machine setup, we utilize eight NVIDIA GeForce RTX 3090 GPUs, each with 24GB of memory, distributed across eight machines connected through 1Gbps networking~\footnote{Note that we purposely configure the inter-machine connection with low networking bandwidth to demonstrate the effect of communication overhead.}. Each machine is equipped with an Intel Xeon Silver 4314 CPU and 256GB of RAM. In the single-machine setup, we further distinguish between the single-GPU mode, using one RTX A6000 GPU, and the single-machine, multi-GPU mode, which defaults to four RTX A6000 GPUs unless specified otherwise. For the multi-machine setup, the default configuration is the multi-machine, multi-GPU mode with eight RTX 3090 GPUs. We use eight NVIDIA RTX A6000 GPUs to test mLoRA's scalability.
\smallskip

\noindent\textbf{Workloads.} In all experiments, we use the natural language generation (NLG) dataset GSM8K~\cite{gsm8k} to evaluate the performance of the training systems. Following the default hyperparameter settings of Alpaca-LoRA~\cite{alpaca_lora}, we fine-tune the PLMs with a batch size of 8, a sequence length of 512, 10 epochs, and a LoRA adapter rank of 16. The LoRA adapter is applied to the linear layers of the PLMs, i.e., $q\_proj$, $k\_proj$, $v\_proj$, and $o\_proj$.
\smallskip

\noindent\textbf{Performance Metrics.} We report the average fine-tuning task completion time, which is the average time required to complete a fine-tuning task, and the system throughput, defined as the total number of tokens the system can train per second.
\smallskip

\noindent\textbf{Baselines.} In the single-GPU environment, we compare mLoRA with HuggingFace \textbf{PEFT}~\cite{peft}, the state-of-the-art library for training parameter-efficient fine-tuning models. 
Due to memory constraints, it is not feasible to use 32fp precision to fine-tune PLMs in this setup (unlike in a multi-GPU setup), so we use 8-bit quantization~\cite{8bitq} and activation checkpointing~\cite{chen2016training} techniques for both mLoRA and PEFT to reduce memory overhead. 

In the multi-GPU environments, whether for single-machine or multiple-machine setups, we compare mLoRA with three state-of-the-art parallelism strategies: 
1) {One Forward Pass followed by One Backward Pass (\textbf{1F1B})}, a synchronous gradient update pipeline parallelism similar to GPipe but more memory-efficient, introduced by PipeDream-Flush~\cite{narayanan2021memory}.
2) {Tensor Parallelism for Transformers (\textbf{TP})}, an optimized model parallelism method for the transformer architecture proposed by Megatron-LM~\cite{megatron-lm};
3) {Fully Sharded Data Parallel ~\cite{fsdp} (\textbf{FSDP})}, an industry-grade parallel LLM training strategy which combines the data and model parallelism and employs the Zero Redundancy Optimizer~\cite{zero, zero-offload} technology proposed by DeepSpeed~\cite{deepspeed}. Note that training LoRA models on multiple GPUs without model parallelism -- where each GPU holds a complete copy of the base model and trains separate LoRA models -- is impractical in our evaluation due to significant memory limitations. Although these constraints can be mitigated by techniques, such as activation checkpointing and 8-bit quantization, they introduce substantial computational overhead and serious precision issues. As a result, we exclude the data parallelism strategy from our multi-GPU environment comparisons.

\subsection{End-to-End Results}
\label{sec:exp_real}
In this section, we present the end-to-end performance results between mLoRA and the state-of-the-art. As the number of simultaneous fine-tuning tasks affects mLoRA's performance, we gradually increase the number of simultaneous fine-tuning tasks until the system's memory capacity is reached.
Each task maintains the parameter settings as outlined in Section ~\ref{sec:exp_setup}, with only modifications to hyperparameters unrelated to throughput, such as learning rate.
\smallskip

\noindent\textbf{Results in single-machine, multi-GPU mode:} As shown in Figure~\ref{fig:end-to-end} (a), (b), and (c), mLoRA achieves an average task completion time reduction of 30\% to 45\% in single-machine, multi-GPU mode with three models of varying parameter scales, thanks to LoRAPP, which decreases communication latency compared to state-of-the-art methods such as TP, and FSDP. Notably, FSDP suffers from additional memory overhead due to parameter replication, preventing it from training a 13B model in a single-machine, multi-GPU mode. In contrast, LoRAPP enables mLoRA to train up to two 13B models simultaneously, as shown in Figure~\ref{fig:end-to-end} (c).
\smallskip

\noindent\textbf{Results in multi-machine, multi-GPU mode:} In the multi-machine, multi-GPU mode, as shown in Figure ~\ref{fig:end-to-end} (d), (e), and (f), it is merely possible for FSDP and TP to train relatively large models given a low-bandwidth cluster (e.g., 1Gpbs in our setup). Although the total GPU memory is the same as that in single-machine, multi-GPU mode, FSDP, and TP incur additional memory overhead on each node, making FSDP impossible to train a 7B or a 13B model and for TP to train a 13B model. In contrast, compared to 1F1B, mLoRA saves 30\% in average task completion time for 7B model, as shown in Figure ~\ref{fig:end-to-end} (e), due to LoRAPP reducing pipeline bubbles. As communication becomes a bottleneck in this setup, resulting in nearly identical training times for both 7B and 13B models.
\smallskip

\noindent\textbf{Results in single GPU:} In the single-GPU setup, as shown in Figure \ref{fig:end-to-end} (g), (h), and (i), ~mLoRA reduces the average task completion time by up to 8\% due to the BatchLoRA operator, which decreases the overhead of launching kernel functions. We note that as the base model size increases, the overhead of launching kernel functions constitutes a smaller proportion, resulting in reduced performance gains by BatchLoRA (e.g., a 2\% reduction with a 13B model).

Moreover, as introduced in Section ~\ref{sec:batchlora}, we can further enhance performance using BatchLoRA when the pipeline reaches zero bubble state (e.g., with more than 4 fine-tuning tasks on 4 GPUs according to Section~\ref{sec:lora_pp_cost}). As shown in Figure ~\ref{fig:end-to-end} (a), when the number of simultaneous training tasks reaches 6, ~mLoRA enables the BatchLoRA operator, resulting in an additional 10\% performance improvement, saving 40\% in average task completion time.
\smallskip

\begin{figure}[t]
    \centering
    \includegraphics[width=0.48\textwidth]{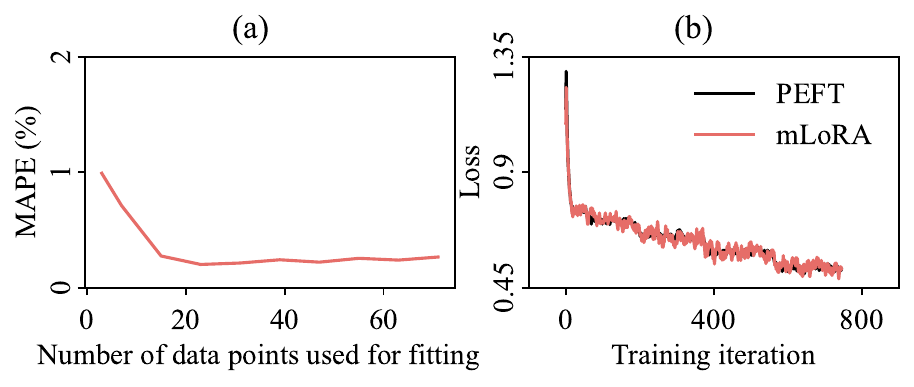}
    \caption{(a) The accuracy of the online model fitting. (b) Model training convergence study.}
    \label{fig:mape-and-loss}
\end{figure}

\noindent\textbf{Model convergence.} We track the loss values for each LoRA adapter during training, as shown in Figure ~\ref{fig:mape-and-loss} (b). ~mLoRA exhibits a convergence trend similar to PEFT, indicating that ~mLoRA achieves the same performance as PEFT.
\smallskip

\noindent\textbf{Memory usage modeling. } We evaluate the accuracy of the online model fitting used in ~mLoRA's scheduler (\S\ref{sec:scheduler}) for predicting the memory usage of each fine-tuning task.
Since the number of data points used to fit the model affects its prediction mean absolute percentage error (MAPE), we use different data points to test its prediction MAPE, as shown in Figure ~\ref{fig:mape-and-loss} (a). The results show that the model achieves a high accuracy, approximately 0.25\% MAPE, even with a limited number of data points. Practically, we can set an error margin of 0.25\% for GPU memory usage estimates to avoid out-of-memory (OOM) errors.

\begin{figure}[t]
    \centering
    \includegraphics[width=0.48\textwidth]{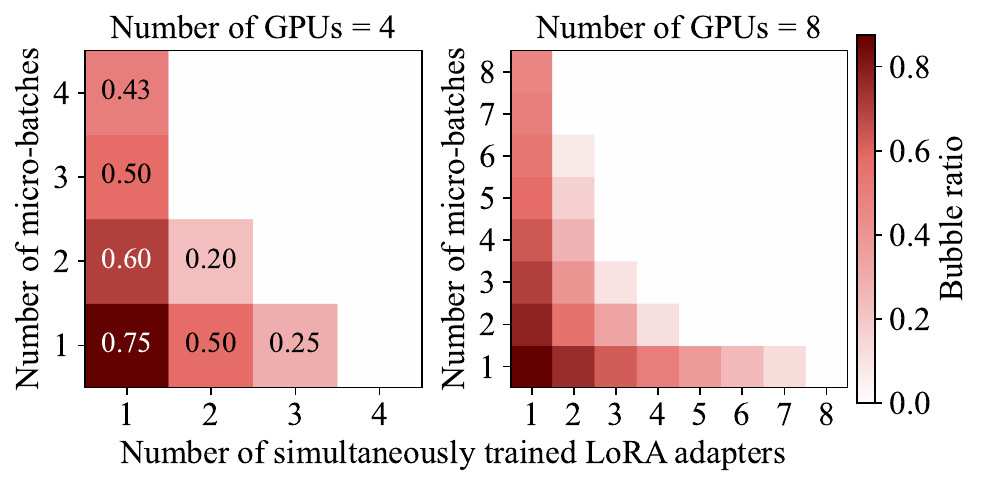}
    \caption{Bubble ratio of LoRAPP.}
    \label{fig:bubble}
\end{figure}

\subsection{Effectiveness of LoRAPP}
\label{sec:exp_lorapp}
In this section, we focus on evaluating the impact of LoRAPP (i.e., no LoRABatch) in single-machine, multi-GPU mode. First, we examine the performance differences between LoRAPP and 1F1B, focusing on bubble ratio. Next, we analyze the differences in communication volume between LoRAPP and TP. We then compare LoRAPP with 1F1B, TP, and FSDP in terms of throughput. Finally, we assess the scalability of mLoRA's LoRAPP parallelism strategy.

\smallskip
\noindent\textbf{Bubble ratio analysis.} Given that the LoRAPP parallelism method also incorporates a mechanism akin to 1F1B, i.e., dividing mini-batch data into micro-batches to reduce the bubble ratio, we analyze the correlation between the bubble ratio, the number of micro-batches, and the number of simultaneously trained LoRA adapters for both mLoRA and 1F1B. 
The results, as depicted in Figure ~\ref{fig:bubble}, show that when fine-tuning a single LoRA adapter, the bubble ratio of mLoRA is comparable to that of 1F1B. 
The bubble ratio of 1F1B decreases gradually with an increasing number of micro-batches but never reaches zero. In contrast, mLoRA can rapidly reduce the bubble ratio to zero by increasing the number of simultaneously trained LoRA adapters, thereby maximizing GPU utilization.

\begin{figure}[t]
    \centering
    \includegraphics[width=0.48\textwidth]{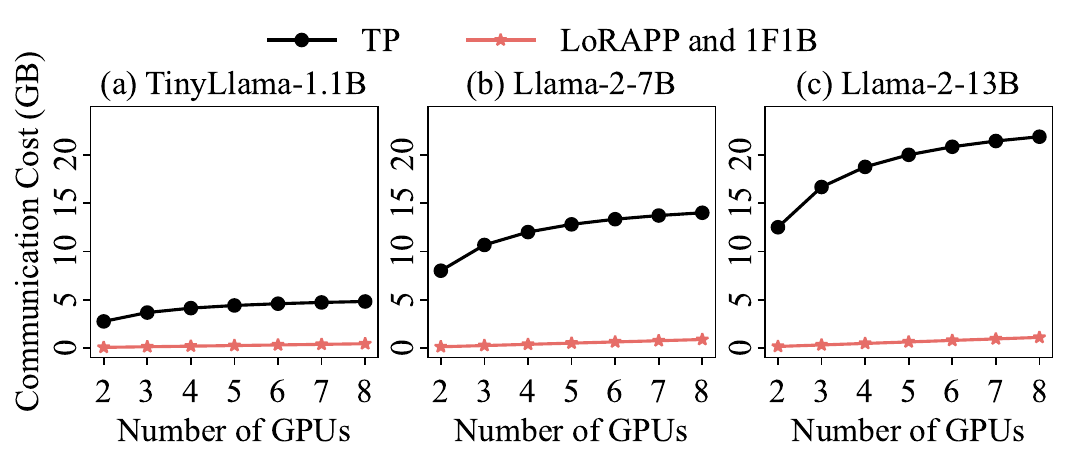}
    \caption{Communication cost comparisons among different parallelism strategies at each training step.}
    \label{fig:pp_cmp_com}
\end{figure}

\smallskip
\noindent\textbf{Communication cost analysis.} 
The communication volume affects the communication time, subsequently impacting the overall training latency or throughput. We measure the communication volume of different parallelism strategies~\footnote{FSDP parallelism strategy encounters OOM errors, so it is omitted in this experiment.}. As shown in Figure ~\ref{fig:pp_cmp_com}, the communication volume for LoRAPP and 1F1B is the same, and significantly smaller than that of TP.

\begin{figure}[t]
    \centering
    \includegraphics[width=0.48\textwidth]{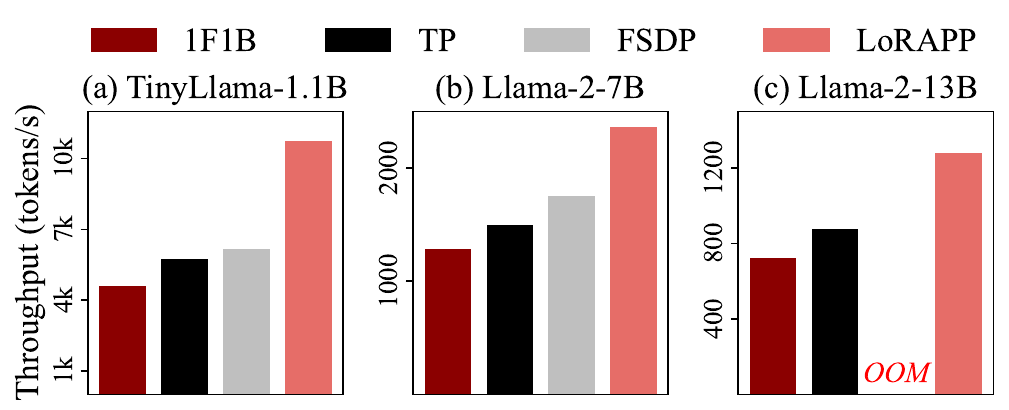}
    \caption{Throughput comparisons among different parallelism strategies.}
    \label{fig:pp_cmp_tp}
\end{figure}

\smallskip
\noindent\textbf{Performace.} 
We first present the highest throughput (i.e., tokens per second) achieved by each approach in Figure ~\ref{fig:pp_cmp_tp}. Due to LoRAPP's smaller communication volume compared to TP (and FSDP) and lower bubble ratio than 1F1B, mLoRA exhibits superior performance. For the 1.1B model, mLoRA's throughput is 75\% higher than FSDP and 86\% higher than TP. For the 7B model, mLoRA outperforms FSDP by 35\% and TP by 58\%. For the 13B model, FSDP encounters an OOM error due to the need for additional memory to store weight copies exceeding GPU capacity, while mLoRA achieves a throughput 46\% higher than TP. As LoRAPP greatly reduces communication overhead, mLoRA benefits the 1.1B model more, which has lower computational overhead and higher communication costs relative to overall training time. In contrast, the 13B model benefits less due to much higher computational overhead.

\begin{figure}[t]
    \centering
    \includegraphics[width=0.48\textwidth]{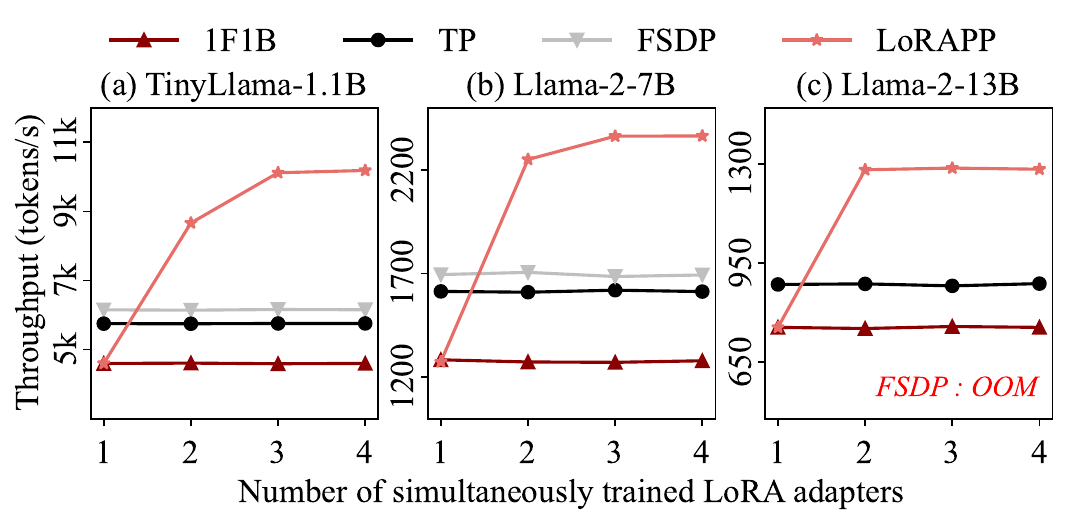}
    \caption{Throughput comparisons among different parallelism strategies with varying numbers of simultaneously trained LoRA adapters.}
    \label{fig:lorapp_task}
\end{figure}

\begin{figure}[t]
    \centering
    \includegraphics[width=0.48\textwidth]{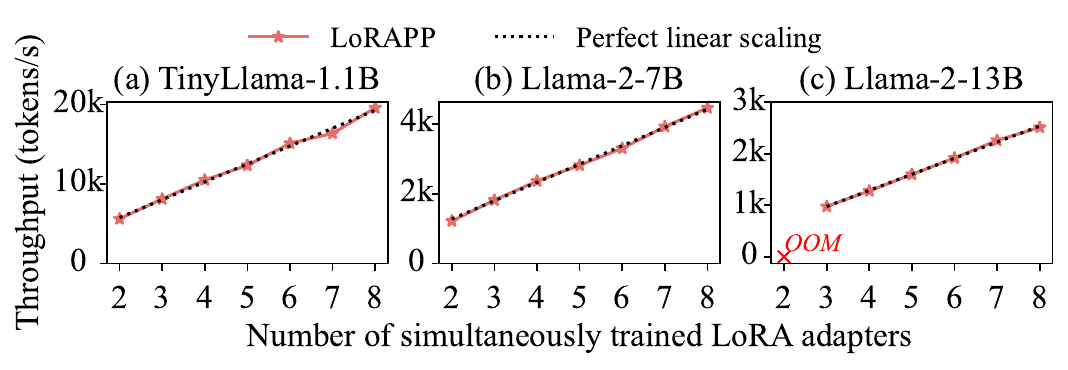}
    \caption{Linear scalability achieved by LoRAPP.}
    \label{fig:scalability}
\end{figure}

We then compare throughput by varying the number of simultaneously trained LoRA adapters. As shown in Figure ~\ref{fig:lorapp_task}, mLoRA's throughput increases with the number of simultaneously trained LoRA adapters until a bubble ratio of zero is achieved (i.e., four LoRA adapters trained simultaneously in a four-GPU setup). Beyond this point, no further improvements are observed, as BatchLoRA is not enabled. In contrast, the throughput of 1F1B, TP, and FSDP remains constant regardless of the number of simultaneously trained LoRA adapters. Specifically, mLoRA outperforms 1F1B in throughput due to its ability to achieve a smaller bubble ratio. Additionally, when only a single LoRA adapter is trained, the higher bubble ratio results in lower throughput for both mLoRA and 1F1B. Furthermore, because mLoRA has a lower communication volume compared to TP and FSDP, it surpasses these strategies in throughput when training more than one LoRA adapter. Note that while launching multiple instances of 1F1B, TP, or FSDP on these GPUs to train multiple adapters could increase throughput, it quickly consumes additional memory and may trigger OOM errors.
\smallskip

\noindent\textbf{Scalability.} To evaluate the scalability of mLoRA, we train LoRA models using an increasing number of GPUs, ranging from 2 to 8. The results, as shown in Figure~\ref{fig:scalability}, indicate that mLoRA's throughput increases linearly with the number of GPUs.

\subsection{Effectiveness of BatchLoRA}
\label{sec:exp_batchlora}
In this section, we examine the impact of the BatchLoRA operator in the single-GPU setup. Given the orthogonal nature of the BatchLoRA operator and the LoRAPP parallelism, the results remain consistent as those in the multi-GPU setup (\S~\ref{sec:exp_real}).

To understand how BatchLoRA mitigates the overhead of kernel function launches, we employ NVIDIA's performance analysis tool, NVIDIA Nsight Systems~\cite{nsys}, to monitor kernel launch times and kernel execution time.
Recall that the effectiveness of the BatchLoRA operator is affected by the number of simultaneously trained LoRA adapters, as it can reduce the number of kernel function launches for multiple tasks to the same level as for a single task. Therefore, we increase the number of simultaneously training LoRA adapters and measure the corresponding kernel launch times and kernel execution times. In addition, to evaluate the effectiveness of mLoRA's graph pruning approach (\S~\ref{sec:graphpruning}) in optimizing the computation graph, we record the forward and backward propagation time and the peak GPU memory consumption.

\begin{figure}[t]
    \centering
    \includegraphics[width=0.49\textwidth]{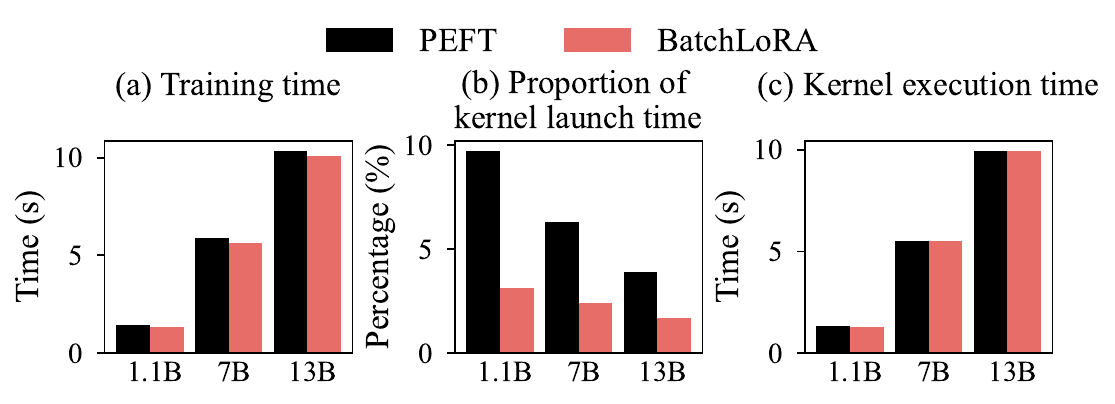}
    \caption{The comparisons of training time, proportion of kernel launch, and kernel execution time between PEFT and BatchLoRA per fine-tuning task at each training step.}
    \label{fig:batchlora_cmp}
\end{figure}

\begin{figure}[t]
    \centering
    \includegraphics[width=0.49\textwidth]{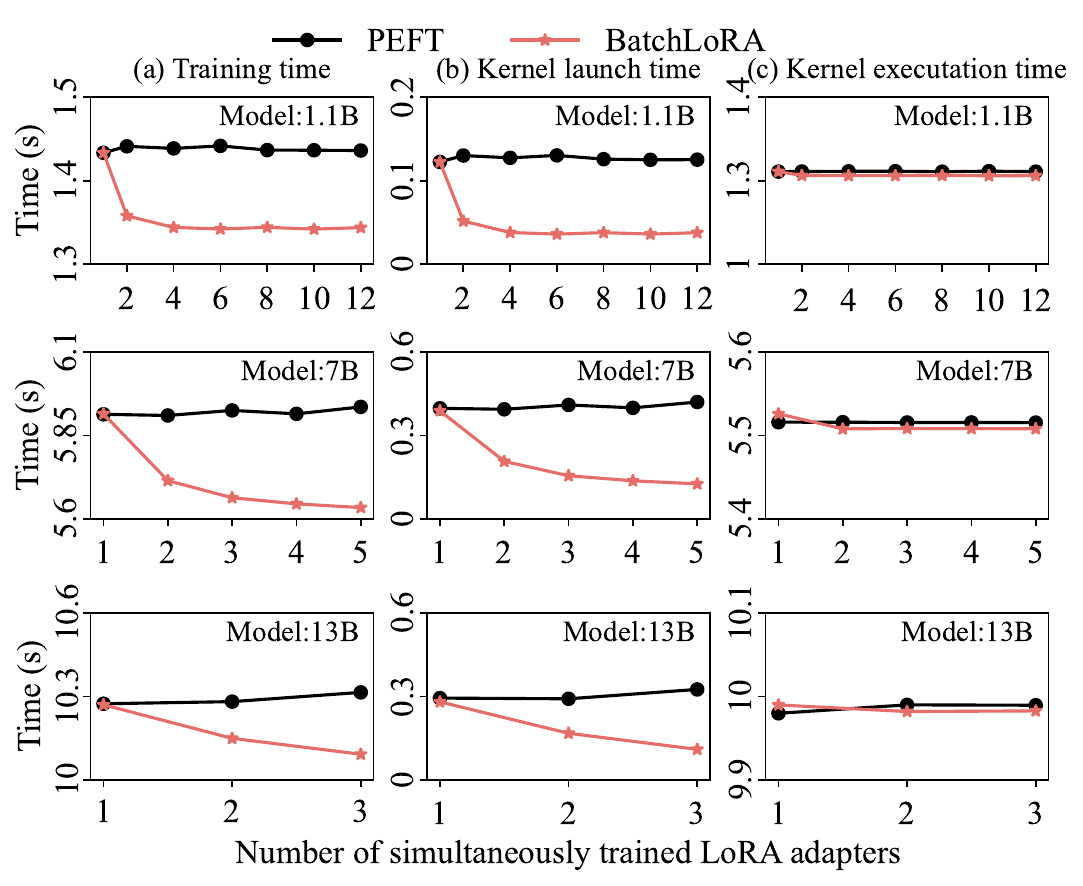}
    \caption{The training time, kernel launch time, and kernel execution time for per fine-tuning task at each training step.}
    \label{fig:batchlora_op_task}
\end{figure}

\begin{figure}[htb]
    \centering
    \includegraphics[width=0.49\textwidth]{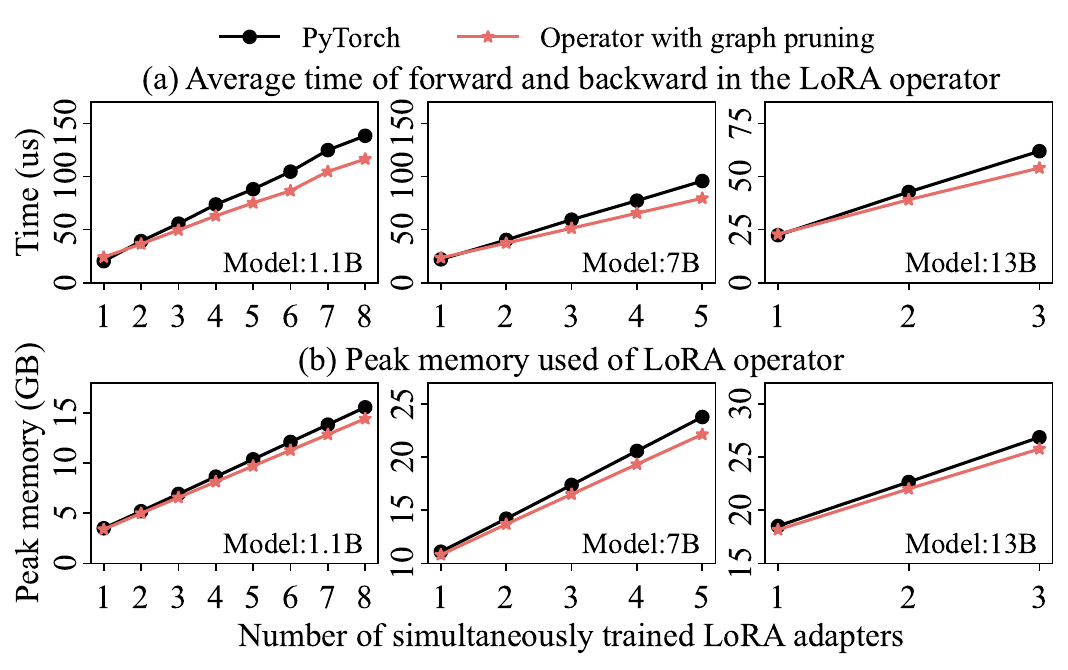}
    \caption{Performance comparisons between the BatchLoRA operator implemented by PyTorch and the operator with graph pruning.}
    \label{fig:batchlora_op_cmp}
\end{figure}

Figure ~\ref{fig:batchlora_cmp} (a) shows that ~mLoRA reduces the training time by 8\% for the 1.1B model, 5\% for the 7B model, and 2\% for the 13B model, compared to HuggingFace PEFT. This improvement is due to the fact that, as illustrated in Figure~\ref{fig:batchlora_cmp} (b), the overhead from launching kernel functions accounts for 10\% of the total overhead for the 1.1B model, and optimizing this aspect leads to significant time savings. In contrast, for the 7B model, the overhead is 7.5\%, and for the 13B model, it is 4\%, resulting in smaller reductions in training time. As shown in Figure~\ref{fig:batchlora_cmp} (c), since the computational workload of the BatchLoRA operator is comparable to that of PEFT, there is almost no difference in kernel execution time.

Figure ~\ref{fig:batchlora_op_task} (a) shows that due to PEFT's inability to batch multiple fine-tuning tasks, the average latency for each training step/iteration remains unchanged. In contrast, with BatchLoRA, the average latency per training step decreases (i.e., becomes more efficient) as the number of simultaneously trained LoRA adapters increases. This is because, as shown in Figure ~\ref{fig:batchlora_op_task} (b) and (c), while the kernel execution time remains nearly identical for both BatchLoRA and PEFT as the number of simultaneously trained LoRA adapters increases, the BatchLoRA operator reduces kernel launch time.

Figure ~\ref{fig:batchlora_op_cmp} shows that mLoRA, through its graph pruning (\S~\ref{sec:graph_pruning}), outperforms the PyTorch-implemented operator in both latency and peak memory usage when simultaneously training multiple fine-tuning tasks. This is because mLoRA's graph pruning reduces memory allocation and copy overhead. Additionally, mLoRA achieves up to a 17\% reduction in latency and a 7\% reduction in peak memory usage as the number of simultaneous fine-tuning tasks increases.

\section{Related Work}
\label{sec:related_work}
\noindent\textbf{Parameter-efficient fine-tuning.} 
Recent works have developed methods for parameter-efficient fine-tuning of large language models. These methods show that fine-tuning is possible with only a small fraction of tuned parameters (e.g. a learnable adapter). The state-of-the-art methods include 1) LoRA-based fine-tuning: LoRA \cite{lora}, AdaLoRA \cite{adalora}, SoRA \cite{sora}, DoRA \cite{dora}, MixLoRA \cite{mixlora}, and MoeLoRA \cite{moelora}; 2) soft prompt-based fine-tuning: Prefix-tuning \cite{prefix-tuning}, prompt-tuning \cite{prompt-tuning}, and P-Tuning \cite{p-tuning, p-tuning-v2}; 3) few-shot fine-tuning: $(IA)^3$ \cite{few-shot}; and 4) selective fine-tuning: diff pruning \cite{diff-pruning} and bitfit \cite{bitfit}. ~mLoRA leverages the plug-and-play feature of LoRA-based fine-tuning to support multi-task training and speed up fine-tuning by pipelining and batching LoRA adapters in concurrent training. While ~mLoRA focuses on a typical LoRA implementation due to its wide adoption, most techniques can be easily applied to other LoRA-based fine-tuning methods as they all follow the same scheme -- i.e., one base model is associated with LoRA adapters.
\smallskip

\noindent\textbf{LoRA-based multi-task systems.} Other works, such as Punica \cite{punica} and S-LoRA \cite{s-lora}, explored the potential of serving multi-task inference services by sharing base-model weights with batched inference requests from different LoRA adapters. However, their optimizations target the inference process, which only involves forward operators. Those optimizations cannot be directly applied to the training process as they do not address redundant operators generated during backward (i.e., split operators).
\smallskip

\noindent\textbf{General-purpose parallelism optimization.} Training LLM with parallelism across devices is a common practice to meet memory demands \cite{dp, fsdp, zero, gpipe, megatron-lm}. There are two paradigms: 1) \textit{Data Parallelism} (DP) \cite{dp}, which distributes a minibatch of data across multiple GPUs; for example, Deepspeed-ZERO \cite{zero} and Pytorch FSDP \cite{fsdp} partition and distribute the model to every GPU for higher memory efficiency. 2) \textit{Model Parallelism} (MP), which allocates subgraphs of a model across different GPUs. The traditional model parallelism methods suffer from high communication traffic and overhead. \textit{Pipeline Parallelism} (PP) and \textit{Tensor Parallelism} (TP) further boost the efficiency of model parallelism. For example, GPipe \cite{gpipe} divides a minibatch into multiple microbatches and injects them into the pipeline, enabling different devices to work with different micro-batches simultaneously. Megatron-LM \cite{megatron-lm} partitions a tensor operation in a layer across GPUs for higher computation and memory efficiency. As discussed in Section~\ref{sec:mm}, existing pipeline parallelism mechanisms remain inefficient due to pipeline bubbles and stalls. In contrast, mLoRA achieves zero bubbles via a LoRA-aware pipeline parallelism scheme. 
\smallskip

\noindent\textbf{Pipeline mechanisms for model training.}
Pipelining has been leveraged to improve the performance
of machine learning systems \cite{pipedream,gpipe,robust_parallel,pipe-SGD,pipelined_back_progation}. Pipelined back propagation \cite{pipelined_back_progation} handles the expensive back propagation. Pipe-SGD pipelines the processing of a minibatch to hide communication time in AllReduce-based systems \cite{pipe-SGD}. A weight prediction technique is proposed to address the staleness issue in pipelined model parallelism \cite{robust_parallel}. PipeDream \cite{pipedream} employs the one-forward-one-backward scheduling algorithm for pipeline execution where the minimum number of mini-batches that is large enough to saturate the pipeline is admitted. ~mLoRA's LoRAPP, specifically targeting LoRA-based fine-tuning, is orthogonal to these optimizations.

\smallskip
\noindent\textbf{GPU kernel launch optimization.} CUDA Graph~\cite{cudagraph} addresses kernel launch overhead by providing a mechanism at the CUDA driver level that allows launching multiple GPU operations with a single CPU operation. Meanwhile, deep learning compilers~\cite{pytorch2, chen2018tvm} mitigate kernel launch overhead at the computational graph level through operator fusion. However, these methods do not support dynamic shapes, and the text data used in fine-tuning often varies in length. mLoRA tackles this issue by consolidating data from multiple fine-tuning tasks into one, reducing the number of operator calls and thus the kernel launch overhead.

\section{Conclusion}
\label{sec:conclusion}
We have presented mLoRA, a fine-tuning system designed and developed for efficiently training multiple LoRA adapters across GPUs and machines. The proposed techniques, including LoRA-aware streamlined pipeline parallelism and a LoRA-efficient training operator, allow mLoRA to fully utilize the computational and memory capacities of a multiple-GPU training cluster. Our extensive evaluation demonstrates that mLoRA significantly reduces average fine-tuning time and improves training throughput compared to state-of-the-art methods.  Deployed in a production environment at AntGroup, mLoRA has achieved over 30\% time savings in selecting optimal hyperparameters for fine-tuning LLM models. Moreover, mLoRA facilitates efficient multi-LoRA fine-tuning on cost-effective GPUs, making LLMs more accessible.

\bibliographystyle{ACM-Reference-Format}
\bibliography{references}

\end{document}